\documentclass[]{catnip}

\usepackage[toc,page,header]{appendix}

\usepackage{minitoc}

\usepackage{url}
\DeclareUnicodeCharacter{1F917}{\textbf{[HF]}}

\definecolor{mygreen}{HTML}{32A626}
\definecolor{myred}{HTML}{E34A17}

\usepackage{wrapfig}
\usepackage{amsmath}
\usepackage{amssymb}
\usepackage{mathtools}
\usepackage{amsthm}
\usepackage{microtype}
\usepackage{titletoc}
\usepackage{bm}
\usepackage{graphicx}
\usepackage{pgfplots}
\pgfplotsset{compat=1.18}
\usepackage{enumitem}
\usepackage{booktabs} 
\usepackage{multirow}
\usepackage{amsmath}
\usepackage{amsthm}
\usepackage{amssymb}
\usepackage{algorithm}
\usepackage{colortbl}
\usepackage{xcolor}         
\usepackage{algorithmic}

\usepackage{color}
\usepackage[table]{xcolor} 
\usepackage{colortbl}      
 
\usepackage{wrapfig}
\usepackage{bbding}
\usepackage[most]{tcolorbox}
\newtcolorbox{response}[1][]{
  colback=gray!5,
  colframe=black,
  fonttitle=\bfseries,
  coltitle=black
  }
\usepackage{longtable}

\definecolor{revc}{HTML}{6D28D9}                 

\definecolor{blockPlan}{HTML}{3F5A77}  
\definecolor{blockCache}{HTML}{4E7C78} 
\definecolor{blockBuf}{HTML}{B07D3F}   

\title{MaineCoon:\\ Pursuing A Real-Time Audio-Visual Social World Model}

\author[]{Catnip AI Team}

\abstract{
As an increasing majority of global video content is consumed on social platforms for interactive social purposes, video generation models built for social worlds are important but largely overlooked by previous studies. In this work, we define the position of social world models and build a prototype model as the first step towards this goal. While previous world models successfully simulate physical environments or gaming world exploration, they remain fundamentally detached from human-centric social dynamics. They typically omit critical auditory information or fail to capture the high-engagement pacing, emotional resonance, and rapid conversational flow that define viral social media. To bridge this gap as the first step to social world models, we present MaineCoon, the first real-time audio-visual autoregressive model that has 22B parameters and is capable of real-time streaming generation and sub-second interaction, with a record-breaking frame rate of up to 47.5 FPS, on a single GPU. To the best of our knowledge, MaineCoon is also the first real-time audio-visual generation model specifically optimized for social-interactive applications. To enable efficient and stable training, we introduce several novel techniques into MaineCoon, including self-resampling, cross-modal representation alignment, domain-aware preference optimization, and reinforced online-policy distillation (ROPD). We also design the first agentic streaming inference framework that supports thousand-second-scale or even longer generation while mitigating drift with agentic cache management and prompt planing. These innovations significantly accelerate training while optimizing real-time inference performance. We believe this work not only sets a new state-of-the-art (SOTA) performance benchmark for high-quality, low-latency, and long-horizon audio-visual autoregressive models, but also points out the paradigm shift desired for next-generation AI-native social platforms.

}

\date{\today}
\project{\url{https://mainecoon.tech/}}

\begin{document}
\maketitle

\section{Introduction}
The rapid progress of generative models has shifted the dominant paradigm from static image synthesis~\citep{esser2024scaling,cai2025z,wu2025qwen} to dynamic video generation~\citep{wan2025wan,wu2025hunyuanvideo,hacohen2026ltx,hong2022cogvideo}, and increasingly toward predictive world models that simulate how a scene evolves over time~\citep{assran2025v,zhu2026sana,tang2025hunyuan,yuan2026fast}. Diffusion transformers (DiT) can now synthesize high-resolution text-to-video and image-to-video clips with strong visual fidelity and coherent motion. 

Nevertheless, conventional video diffusion models~\citep{wan2025wan,wu2025hunyuanvideo,hacohen2026ltx,hong2022cogvideo} remain constrained by two fundamental limitations. First, they are slow and expensive. Generating a video requires repeatedly processing a large number of spatiotemporal tokens over many denoising steps, while the computational and memory costs of full self-attention grow rapidly with video length and resolution. A large body of works have therefore explored efficient video generation \citep{shao2026efficient} through four major directions, including step distillation~\citep{yin2024improved,yin2024one,shao2025magicdistillation}, efficient attention~\citep{zhang2025spargeattention,zhang2026faster,li2026pisa,shao2026liveditor}, model compression~\citep{zhao2024vidit,chen2025qdit,shao2026fastlightgen}, and cache optimization~\citep{liu2025timestep}. These techniques reduce the number of denoising steps~\citep{shao2025magicdistillation,bai2026optimizing,yin2024improved,yang2026sparse} or the computational cost of each model evaluation~\citep{zhang2026faster,li2026pisa,shao2026fastlightgen}, substantially improving offline generation efficiency. However, they do not fully eliminate the first limitation, because the underlying self-attention computation still becomes increasingly expensive as the spatiotemporal context grows. Second, conventional video diffusion models are not real-time. They typically denoise an entire clip jointly under bidirectional temporal attention, so intermediate frames cannot be finalized and emitted while generation is still in progress. Consequently, efficiency-oriented techniques may accelerate the production of a complete video, but they do not fundamentally enable causal streaming generation or continuous interaction with users.

To fundamentally overcome the slow and non-streaming nature of bidirectional video diffusion models, recent works have explored streaming and autoregressive video generation~\citep{yin2025slow,huang2026self,yang2025longlive,liu2025rolling}. These methods generate frames or short chunks sequentially, reuse historical states through key-value caching, and often reduce sampling to only a few steps, enabling low-latency interactive video synthesis. However, existing streaming video generation models at least suffer from three unsatisfactory limitations. First, they~\citep{krea_realtime_14b,yuan2026helios,zhu2026causal} mainly focused on visual generation and either omit audio or use it merely as a condition, preventing the joint generation of speech, ambient sound, facial motion, and their temporal synchronization. Second, long autoregressive rollouts also remain vulnerable to error accumulation and temporal drift~\citep{su2026omniforcing,huang2026self}. Minute-scale audio-visual generation is still an open challenge so far. Third, while some of them claimed real-time generation speed, they cannot achieve real-time generation on a single GPU, which means they are still too expensive to support large-scale social-interactive applications. Consequently, none of them can yet deliver a long-horizon scalable real-time social audio-visual experience.

These limitations are particularly consequential for social video, whose appeal largely depends on human expression, speech rhythm, emotional resonance, conversational pacing, and rapid audience feedback. The central challenge is therefore no longer only to generate visually compelling videos, but to build real-time audio-visual generators that can participate in social interaction with low latency, long-horizon stability, and synchronized multimodal behavior.

\paragraph{Social World Model} Interestingly, most video contents in the world are actually watched on social platforms and created for interactive social purposes. The ratio is even continually increasing. While recent interactive world models have made progress in simulating physical environments and gaming world exploration, they remain fundamentally detached from the most consumed video content and human-centric interactive social purposes. To bridge this gap, we define the position of \textit{social world models}, which is a generative paradigm engineered to understand, simulate, react to human-centric social dynamics. Human social interaction has its own ``social physics'' which include a set of highly structured, multi-modal behavioral rules that a model can learn auto-regressively. Social dynamics Unlike traditional world models that predict environmental transitions or object trajectories, \textit{a social world model learns to actively observe users, internally simulate social dynamics, and react to users in real time. }This will require capturing the main attractions of social videos, including rich human expression, tight audio-lip-motion synchronization, rapid conversation, high-engagement pacing, emotional resonance, and reaction prediction across extended, minute-level horizons. The position of social world models is largely overlooked by previous works. Social world models will shift Gen AI from a passive content-generation tool into an active, responsive participant in the human social fabric. It will be difficult to see successful next-generation AI-native social platforms without social world models that enable real-time human-centric social-interactive generation.

The full realization of social world models requires multiple capabilities, including active multimodal observation, internal user-state simulation, memory, planning, and real-time audio-visual generation. In this technical report, we focus on the generative core of this system: whether a large audio-visual model can generate social video in a native streaming autoregressive manner, maintain synchronized speech and visual motion, support long-horizon continuation, and still run in real time on a single GPU. This generative core is a necessary first step toward practical social world models, because higher-level observation and social simulation modules ultimately need to be grounded in a low-latency audio-visual interface. 

We believe that the first step towards the paradigm of social world models requires a strong, ultra-low-latency, real-time social video generation model. To bridge this gap, we propose \textit{MaineCoon}, a 22B-parameter real-time social audio-visual autoregressive model. MaineCoon generates synchronized audio and video chunk by chunk under a causal streaming regime, supports sub-second interaction. Unlike offline bidirectional video generators, MaineCoon is designed around deployment-time streaming from the beginning: its data infra, training framework, attention pattern, context distribution, KV-cache usage, and agentic streaming inference are all optimized for real-time social audio-visual generation.

\paragraph{Contributions} We mainly made five contributions.

\textbf{First}, we build a 22B real-time interactive audio-visual autoregressive model capable of streaming generation and sub-second interaction, with a record-breaking frame rate of up to 47.5 FPS, on a single H100 GPU. Audio-visual generation cost drops significantly below \$0.001 per second and continues to fall. 

\textbf{Second}, we position the paradigm of \textit{social world models} for social-interactive purposes. MaineCoon serves as the first generative core towards this paradigm and provides a technical foundation for next-generation AI-native social platforms.

\textbf{Third}, we propose a multi-stage forcing-free streaming training paradigm that includes self-resampling, cross-modal representation alignment, domain-aware preference optimization, and reinforced online-policy distillation (ROPD). These components enables 22B-scale native and efficient streaming audio-visual training.

\textbf{Fourth}, we design an agentic streaming inference framework that supports thousand-second-scale or even longer generation while mitigating drift through agentic cache management, chunk commitment, long-context rollout, and prompt planning.

\textbf{Fifth}, we introduce \textit{SocialVideo Bench}, a benchmark focused on audio-visual generation for social videos, and demonstrate the SOTA performance of MaineCoon. With 9 representative evaluation metrics, SocialVideo Bench provides a comprehensive evaluation of visual quality, motion, audio quality, audio-visual alignment, and social-video harmony. Experiments show that MaineCoon significantly outperforms 7 representative open audio-visual generation models while achieving the fastest generation speed, setting a new state of the art for real-time social video generation.

\section{Data}

In this section, we present the social-video data infrastructure behind MaineCoon.

Four features set this infrastructure apart from a generic video-data pipeline. First, it is built on social-oriented raw data: short social videos, valued for their liveness, rather than cinematic footage. Second, it judges every clip on both video quality and audio-visual quality, since a social world model must render speech, lip motion, and expression in tight synchrony. Third, its filtering and standardization are designed to yield data suited to audio-visual streaming training, including the long-horizon, prompt-switching supervision that streaming requires. Fourth, the infrastructure is itself an efficient streaming system: raw short videos are continuously ingested from social-media platforms, passed through a cascade of automatic filters, and converted into standardized, training-ready audio-visual clips at a throughput of up to one hundred thousand videos per day. We first articulate what distinguishes a good social video from a conventional cinematic one in \cref{subsec:good_social_video}, then describe the pipeline that curates such videos at scale in \cref{subsec:data_pipeline}.

\subsection{What Is A Good Social Video?} \label{subsec:good_social_video}

The video contents popular on social platforms essentially differs from the cinematic footages that most existing generative video models are trained to imitate. Cinematic video is authored: it is staged, lit, and edited to serve a director's narrative, so its visual language favours composition, scale, and spectacle. A social video, by contrast, is valued for what we call its liveness: the immediate impression that a real person is present, attending to the viewer, and reacting in the moment. This impression is carried less by scene grandeur than by fine-grained human signals: the direction and steadiness of gaze, micro-expressions, head and hand gestures, the timing and prosody of speech, and audible emotion.

Accordingly, we take a good social video to be one that is natural and human-centric rather than staged, that sustains tight synchronization among speech, lip motion, and facial expression, and that remains temporally consistent and coherent over its duration. These are precisely the properties a social world model must learn to generate, and they are abundant, though heavily entangled with noise, in the short videos that dominate social-media platforms. We therefore build our training dataset primarily from a massive pool of such short videos, and design the pipeline below to isolate, at scale, the clips that exhibit these properties at high quality.

\subsection{Data Pipeline} \label{subsec:data_pipeline}

Our data pipeline draws on two complementary sources: synthetic audio-visual video, generated by a strong teacher model, and real social video, curated from social-media short videos.

We first describe the synthetic source because it is built to achieve stable early streaming training and also contribute to preference data pairs for post-training. Real social videos supply the main body of the dataset. Both sources are reduced to a common, training-ready representation, standardized clips together with their text, audio, and visual conditioning, and merged into a set of domain-balanced training buckets optimized for social purposes. Every stage is stateless at the clip level and sharded across many GPU workers. In the automatic data pipeline, generation, filtering, and feature extraction proceed concurrently rather than as discrete offline passes.

Each source is curated by its own pipeline. The synthetic source proceeds through scenario planning, segmented generation, quality filtering, and training-ready packaging; the real source runs a cascade of low-level filtering, high-level filtering, speech transcription, and encoding. The curated real clips are then clustered by framing and motion into domains and balanced into training buckets, up-weighting the rare domains against the dominant talking-head bulk. The subsections below describe each in turn.

\subsubsection{Synthetic Data Pipeline}

\begin{figure}[t]
    \centering
    \IfFileExists{Assets/Figures/data-pipeline-fig-synth.pdf}{%
      \includegraphics[width=\linewidth]{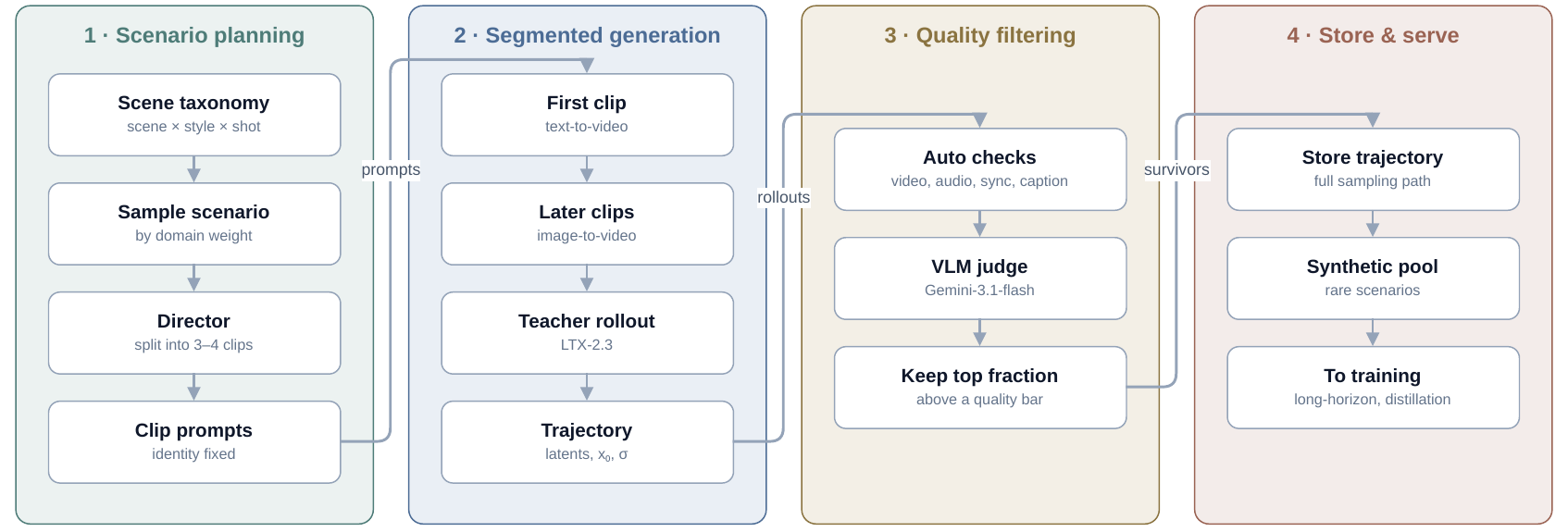}}{%
      \fbox{\parbox[c][2.4cm][c]{0.985\linewidth}{\centering\small\itshape Figure pending: export \texttt{data-pipeline-fig-synth.pdf} from the synthetic tab of \texttt{Assets/Figures/data-pipeline.drawio}.}}}
    \caption{\textbf{Synthetic audio-visual data pipeline.} A director-style language model samples a scenario from a scene, style, and shot taxonomy and splits it into three to four linked clips that hold character identity fixed while the shot, action, dialogue, and sound advance. The first clip is generated text-to-video and each later clip image-to-video from the preceding final frame, so one rollout spans several prompt switches. A multi-stage quality gate keeps only high-quality clips, and the full sampling trajectory of every survivor is stored for the consistency and step-distillation objectives, not only the final video. }
    \label{fig:data_synthetic}
\end{figure}

We synthesize short audio-visual clips with the teacher model LTX-2.3 for two reasons specific to the streaming training of \cref{sec:training}, not merely to add volume. First, the teacher generates fixed-length clips offline, and converting it into a real-time streaming model requires segmented, prompt-switching supervision that raw social video does not provide; we generate this data so the streaming conversion has matched long-horizon targets to learn from. Second, we want the model to switch and inject prompts on the fly across a long streaming rollout, so we produce multi-segment stories whose prompt changes from segment to segment, supplying the long-horizon, prompt-switching supervision the agentic streaming inference of \cref{sec:inference} depends on. Because we generate the clips ourselves, we also retain each one's full sampling trajectory, which the consistency and step-distillation objectives of \cref{subsec:native_sr} reuse directly. In~\cref{fig:data_synthetic} we summarize this pipeline.

\paragraph{Scenario planning} We first plan a multi-clip story whose prompt switches over a fixed subject. To keep the synthetic corpus diverse rather than repetitive, so the teacher is exercised across the full breadth of social-video situations instead of a few recurring setups, we draw the prompts from a director-style language-model pipeline over a broad taxonomy of 225 scenes across ten thematic groups, 15 visual styles, and 12 camera shots: each scenario is sampled by a domain weight that favors conversational and performance scenes and decomposed into three to four consecutive clips that together tell a roughly twenty-second story, establishing the scene, developing it, and closing on a reaction. Across the clips the character's identity and appearance stay fixed while the shot, action, dialogue, and sound advance, so consecutive clips carry different prompts over the same subject. Each clip prompt is written in the present tense, names its visual style, interleaves sound with action, quotes the spoken lines, and is constrained to contain no on-screen text or graphics so the teacher never learns to render them.

\paragraph{Segmented generation} We then render the planned story with the teacher into the long-horizon, prompt-switching rollouts the streaming model must learn to extrapolate: the first clip is generated text-to-video, and each subsequent clip image-to-video seeded by the previous clip's final frame, as fixed-length segments of roughly five seconds at one of a few resolution and frame-count buckets the teacher supports.

\paragraph{Quality filtering} Generated clips pass a composite quality gate before entering the dataset. A clip is rejected outright if its video fails to decode or carries no audio. The rest are scored over uniformly sampled frames and the full audio by four analyzers: video quality from frame rate, resolution, black and static frames, and scene cuts; audio quality from signal-to-noise ratio, silence, clipping, and a Silero voice-activity~\citep{silerovad} label of speech, music, ambient, or noise; audio-visual sync from stream-duration agreement and the correlation between visual motion and audio energy; and caption quality from length, lexical diversity, repetition, structured-field completeness, and agreement with the detected audio category. A vision-language judge, Gemini-3.1-flash~\citep{team2023gemini}, then rates the consistency between uniformly sampled frames and the caption, contributing a fifth score. The scores are combined into a weighted composite, and only clips above a quality bar survive, so that a small fraction of generations enter the dataset.

\paragraph{Training-ready packaging} A final post-processing step turns every survivor into a training-ready record. Rather than keeping only the final video, we cache the full sampling trajectory: the initial noise, the per-step latents and clean-signal $\mathbf{x}_0$ predictions along the ODE, the noise schedule, and the text conditioning for both modalities. The packaged clip is then ready for training.

\subsubsection{Real Data Pipeline}

\begin{figure}[t]
    \centering
    \IfFileExists{Assets/Figures/data-pipeline-fig1.pdf}{%
      \includegraphics[width=\linewidth]{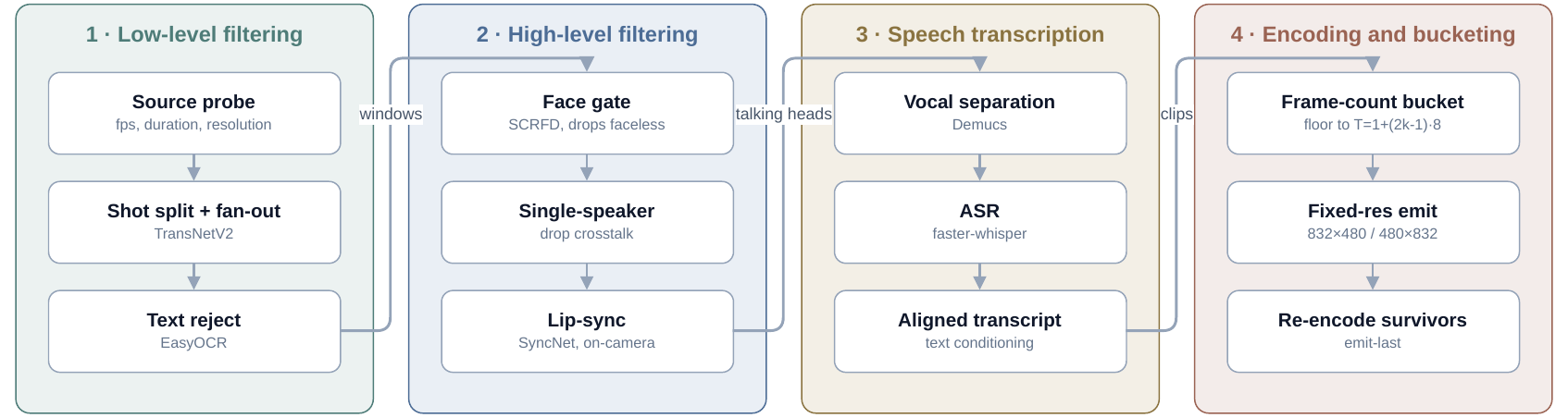}}{%
      \fbox{\parbox[c][2.4cm][c]{0.985\linewidth}{\centering\small\itshape Figure pending: export \texttt{data-pipeline-fig1.pdf} from tab~1 of \texttt{Assets/Figures/data-pipeline.drawio}.}}}
    \caption{\textbf{Real-video data pipeline.} A speech-strict, person-centric cascade curates training clips from raw social-media short videos. Low-level filtering probes each source, splits it into single-shot windows, and drops windows with persistent on-screen text. High-level filtering keeps only talking-head windows: an SCRFD face detector discards faceless content, an overlapping-speech check removes multi-speaker crosstalk, and SyncNet lip-sync verification keeps windows whose visible speaker produces the audio. Demucs and faster-whisper then transcribe the speech into time-aligned text conditioning, and surviving windows are bucketed to the generator's frame grid and re-encoded at fixed resolutions. Domain clustering and balancing is a separate step and is not shown.}
    \label{fig:data_primary}
\end{figure}

Real videos supply the bulk of the dataset. We curate them from tens of millions of raw social-media videos with a speech-strict, person-centric cascade that keeps only single-shot windows showing a clearly visible person speaking on camera. The cheapest rejections run first, so the costly speech and lip-sync models process only survivors. A summary of the cascade is provided in~\cref{fig:data_primary}.

\paragraph{Low-level filtering} The aim of the first stage is to spend almost no compute on footage that cannot qualify and to reduce every source to single continuous shots, since the model trains on single-shot windows and a cut inside a clip would teach it spurious scene changes. A probe screens each source and rejects any whose frame rate falls outside a normal 23 to 32 FPS range or whose duration or resolution is unusable. TransNetV2~\citep{soucek2020transnetv2} then segments the survivors into shots of three to twenty seconds, which are fanned out into per-shot windows, and EasyOCR~\citep{easyocr} drops any window carrying persistent on-screen text such as subtitles, captions, or watermarks, which the generator must never learn to render.

\paragraph{High-level filtering} This stage enforces the property that defines the corpus, a single clearly visible person speaking on camera, and it runs the cascade's heaviest models, so its checks are ordered cheapest and most discriminative first. An SCRFD~\citep{guo2022scrfd} face detector is applied first, directly on the raw source, and is the single most effective filter: it discards close to half of all candidates that contain no detectable face before any costlier model runs, and is re-applied per window after shot segmentation. On the windows that remain, an overlapping-speech check rejects clips in which two or more people speak at once, using a lightweight spectral pitch heuristic rather than a heavy diarization model, so every kept clip carries clean single-speaker audio. SyncNet~\citep{chung2016syncnet} then verifies lip-sync by correlating mouth motion with the audio within a tolerance of a few frames, keeping only windows whose visible speaker is the source of the speech. Animals, scenery, and off-screen narration fall away by construction, leaving the human-centric, speech-synchronized clips of \cref{subsec:good_social_video}.

\paragraph{Speech transcription} The windows that survive already show a single visible speaker in sync with the audio; this stage converts that speech into the text the model is conditioned on. Demucs~\citep{defossez2019demucs} first separates the vocal track from background music and ambient noise so the recognizer sees clean speech, and faster-whisper~\citep{radford2023whisper} then performs automatic speech recognition, transcribing the vocals into sentence-level segments with start and end time. The resulting time-aligned transcript is paired with the clip as its text conditioning.

\paragraph{Encoding and bucketing} The final stage standardizes every survivor to the fixed temporal grid and resolution the generator trains on, so a clip can be loaded for training without further reshaping. The generator accepts only frame counts that land on its latent temporal grid, those of the form $T = 1 + (2k-1)\cdot 8$ for integer $k \ge 1$, that is $T \in \{9, 25, 41, 57, \dots\}$, one anchor frame on top of an odd multiple of the temporal compression factor $8$ of the latent video tokenizer. A shot of $D$ frames is mapped to the largest admissible bucket $T \le D$, flooring rather than rounding so the emitted window stays strictly inside the detected shot; rounding up would read past the cut and splice in frames from the next shot. Each window is then scaled to a short edge of $480$ and center-cropped to one of two fixed resolutions, $832\times480$ for landscape and $480\times832$ for portrait, and re-encoded with libx264. The cascade is emit-last: it reads and re-encodes the source only for windows that pass every preceding filter.

\subsubsection{Post-Training Data Pipeline}

The curated real corpus is built for pre-training, but post-training needs something more targeted, namely data drawn from the specific domains where the model is weakest. The corpus is far from uniform: it is dominated by easy close-up, low-motion, talking-head footage, while the wide-shot, high-motion, and multi-person clips the model finds hardest are rare, so training on it as-is lets the dominant bulk drown out exactly the domains we want to reinforce. We therefore run a second classification pass over the already-filtered clips, labeling each by its visual domain, and from these labels select a domain-balanced set that up-weights the hard domains.

\paragraph{Domain profiling} A single pass over each clip decodes eight uniformly sampled frames and scores them along two complementary axes. Shot scale is the fraction of the frame occupied by the largest person, detected with YOLO11x~\citep{yolo11}, separating wide shots from close-ups; motion is the mean change between consecutive SigLIP-SO400M~\citep{zhai2023siglip} frame embeddings, separating dynamic footage from near-static talking heads. An offline analysis then thresholds the two scores into a grid of shot scale and motion, cross-tabulates that grid against content clusters of the same embeddings, and draws per-quadrant samples for audit.

\paragraph{Domain-balanced selection} We group the clips into a handful of training domains, among them wide shot, high motion, multi-person, first-person, close-up, and medium shot. We up-weight the rare, demanding domains and set aside the dominant intersection of close framing and low motion, the static talking-head bulk that otherwise forms the large majority of the corpus. The resulting domain-balanced set is the data we reserve for post-training, weighted toward the hardest domains such as wide-shot and high-motion social video.

\subsection{SocialVideo Bench}
In this subsection, we introduce \textit{SocialVideo Bench}, a domain-focused benchmark for evaluating audio-visual generation on representative social-video contents.

\begin{table}[h]
\centering
\caption{\textbf{Domain composition of SocialVideo-Bench.}
SocialVideo-Bench contains 700 samples evenly distributed across seven representative social-video domains. We obtain these domains from a content clustering of social videos: clustering clips on their visual-content embeddings surfaces the recurring content groups of social media, which we distill into the seven evaluation domains.}
\label{tab:bench}
\resizebox{\columnwidth}{!}{%
\begin{tabular}{lc|l}
\toprule
Domain & Samples & Representative Content \\
\midrule
Dense Speech & 100 & Continuous speech, narration, explanation, and monologue \\
Two-Person Interaction & 100 & Conversations, interviews, debates, and interpersonal reactions \\
Music and Vocal & 100 & Singing, vocal performance, and music-centered human activities \\
Emotional Performance & 100 & Expressive speech, facial emotion, and emotion-driven behavior \\
Dance & 100 & Solo or multi-person dance with prominent rhythmic body motion \\
Creative Stress Test & 100 & Creative combinations of human actions, audio events, and complex scenes \\
Social Memes & 100 & Social content involving humor, reactions, and narrative reversals \\
\midrule
\textbf{Total} & \textbf{700} & Seven balanced social-video domains \\
\bottomrule
\end{tabular}%
}
\end{table}

Existing video-generation benchmarks~\citep{huang2024vbench,liu2025javisdit,wu2024longvideobench} provide broad coverage of visual fidelity, motion quality, and text-video alignment, but their prompt distributions are typically dominated by general scenes, object motion, cinematic content, or physical interactions. As a result, they provide limited coverage of the content domains that are most representative of social video, where humans, speech, music, expression, and interpersonal interaction occupy a central role.

We therefore introduce \textit{SocialVideo Bench}, a domain-focused benchmark for evaluating audio-visual generation on representative social-video content. Rather than introducing a new set of evaluation metrics, SocialVideo Bench adopts established metrics for visual quality, motion quality, audio quality, text-video alignment, and audio-visual alignment, while constructing a balanced evaluation set across seven social-video categories. The benchmark contains 700 prompt pairs, with 100 pairs in each category. Every sample consists of two consecutive 10-second segments, with the prompt updated at the 10-second boundary. This setting evaluates generation quality on social-video domains as well as the model's ability to follow a changing prompt while maintaining temporal and audio-visual consistency.

\section{Training} \label{sec:training}

In this section, we present a novel forcing-free training framework for MaineCoon. MaineCoon is a native streaming autoregressive audio–visual generator, optimized directly under the same chunk-by-chunk causal regime used at inference instead of be not distilled from a non-causal teacher via teacher forcing. 

\begin{figure}
    \centering
    \includegraphics[width=1.0\linewidth]{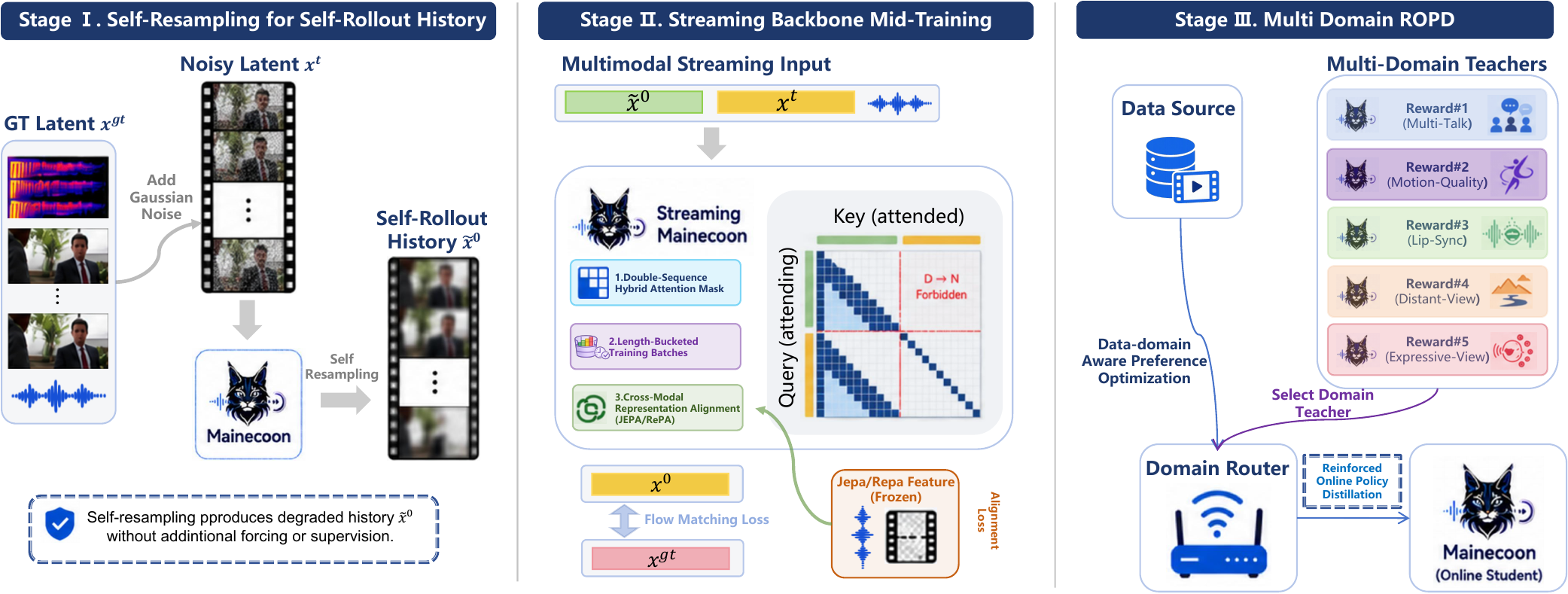}
    \caption{Overview of the MaineCoon training framework. 
    The training pipeline integrates three key techniques: 
    (i)~native streaming AR with self‑resampling, (ii)~representation alignment, (iii)~domain‑aware preference optimization followed by ROPD consolidation.
    These components together enable a 22B causal audio‑visual generator to be trained on less than one million clips within 10k GPU hours while supporting real‑time streaming inference.}
    \label{fig:Train_Framework}
\end{figure}

The training recipe consists of four parts: (1)~native streaming AR training with self-resampling, which teaches the model to generate each chunk causally while conditioning on its own degraded history (\cref{subsec:native_sr}); (2)~representation alignment, which accelerates training and strengthens audio-visual correspondence through pretrained-encoder supervision (\cref{subsec:repa}); (3)~domain-aware preference optimization and reinforced online-policy distillation, which adapt the model to heterogeneous social-video domains and consolidate them into a single deployable streaming policy (\cref{subsec:dpo_ropd}); and (4)~the training infrastructure that makes streaming training efficient and practical (\cref{subsec:training_infra}). An overview of our training framework is illustrated in \cref{fig:Train_Framework}.

\subsection{Native Streaming Autoregressive Training with Self-Resampling} \label{subsec:native_sr}

Our design is inspired by the resampling forcing~\citep{guo2025end}, which reduces the train-test context gap in
autoregressive diffusion generation by exposing the model to degraded histories
induced by its own predictions. We adapt this idea to synchronized
audio-visual streaming generation. Instead of distilling streaming behavior
from a future-aware non-causal teacher or relying on a forcing-style
distillation objective, MaineCoon is post-trained directly under the same
chunk-by-chunk causal rollout regime used at inference. The key idea is to train
each target audio-visual chunk not only from clean ground-truth histories, but
also from model-induced degraded histories constructed through stop-gradient
self-resampling. This gives the model robustness to the imperfect contexts it
will encounter during long-horizon streaming deployment.

\paragraph{Native streaming AR training} We treat the audio-visual signal as
a sequence of synchronized chunks,
\begin{equation}
\mathbf{x}_{1:T}=\{\mathbf{x}_1,\ldots,\mathbf{x}_T\},
\qquad
\mathbf{x}_t=(\mathbf{x}_t^{v},\mathbf{x}_t^{a}),
\end{equation}
where $\mathbf{x}_t^{v}$ and $\mathbf{x}_t^{a}$ are the visual and audio latents
of chunk $t$. Here, we set the training chunk size to 2, where each chunk is a paired tuple of two latents sharing the same temporal extent. And $T$ is the number of chunks.
We model the sequence with a causal autoregressive factorization,
\begin{equation}
p_\theta(\mathbf{x}_{1:T}\mid\mathbf{c})
=\prod_{t=1}^{T} p_\theta(\mathbf{x}_t\mid\mathbf{x}_{<t},\mathbf{c}),
\end{equation}
where $\mathbf{x}_{<t}=(\mathbf{x}_1,\ldots,\mathbf{x}_{t-1})$ are the committed
past chunks and $\mathbf{c}$ collects the available conditions (text, speech,
scene, and social-domain signals); each chunk $t$ may attend only to
$\mathbf{x}_{<t}$ and never to future audio or video.

Each chunk is generated by a conditional flow-matching (rectified-flow) process.
For a clean target chunk $\mathbf{x}_t$, a noise level $\tau\in[0,1]$, and
Gaussian noise $\boldsymbol{\epsilon}\sim\mathcal{N}(\mathbf{0},\mathbf{I})$, we
define the noised chunk and its velocity target as
\begin{equation}
\mathbf{z}_{t,\tau}=(1-\tau)\,\mathbf{x}_t+\tau\,\boldsymbol{\epsilon},
\qquad
\mathbf{u}_{t,\tau}=\boldsymbol{\epsilon}-\mathbf{x}_t .
\end{equation}
The network $f_\theta$ is a causal audio-visual diffusion transformer that
takes a noised target chunk, its noise level, the committed past chunks, and the
conditions, and predicts the velocity field; it is trained by the
denoising objective
\begin{equation}
\mathcal{L}_{\mathrm{AR}}
=\mathbb{E}_{t,\tau,\boldsymbol{\epsilon}}\!\left[
\left\| f_\theta\!\left(\mathbf{z}_{t,\tau},\tau,\mathbf{x}_{<t},\mathbf{c}\right)
-\mathbf{u}_{t,\tau}\right\|_2^2
\right].
\end{equation}
At this stage the historical context $\mathbf{x}_{<t}$ is drawn from real data,
so the objective resembles teacher-forced autoregressive training; crucially,
however, its attention mask, chunk ordering, and KV-cache update pattern are
already identical to those of streaming inference.

\begin{figure}[t!]
    \centering
    \begin{minipage}[t]{0.435\textwidth}
        \centering
        \begin{subfigure}[t]{\linewidth}
            \centering
            \includegraphics[width=\linewidth]{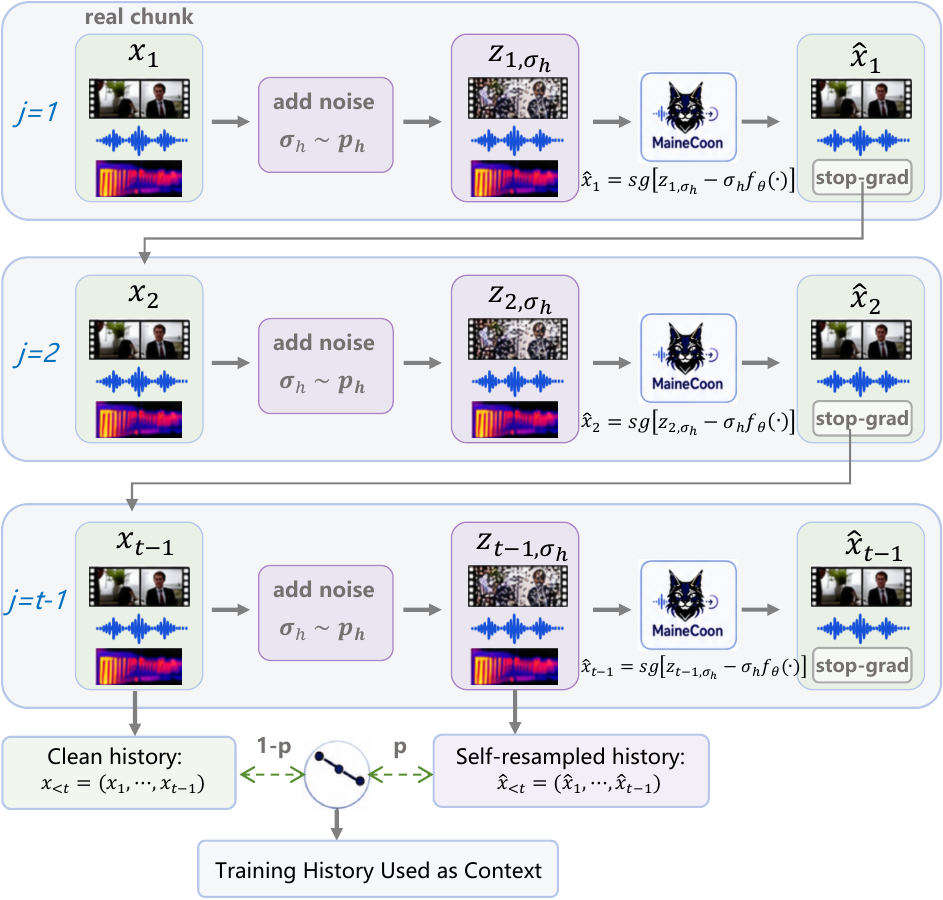}
            \caption{
                Audio-visual self-resampling constructs model-induced degraded
                histories by perturbing real past chunks and denoising them under
                stop-gradient.
            }
            \label{fig:sr_sampling}
        \end{subfigure}
    \end{minipage}
    \hfill
    \begin{minipage}[t]{0.535\textwidth}
        \centering
        \begin{subfigure}[t]{\linewidth}
            \centering
            \includegraphics[width=\linewidth]{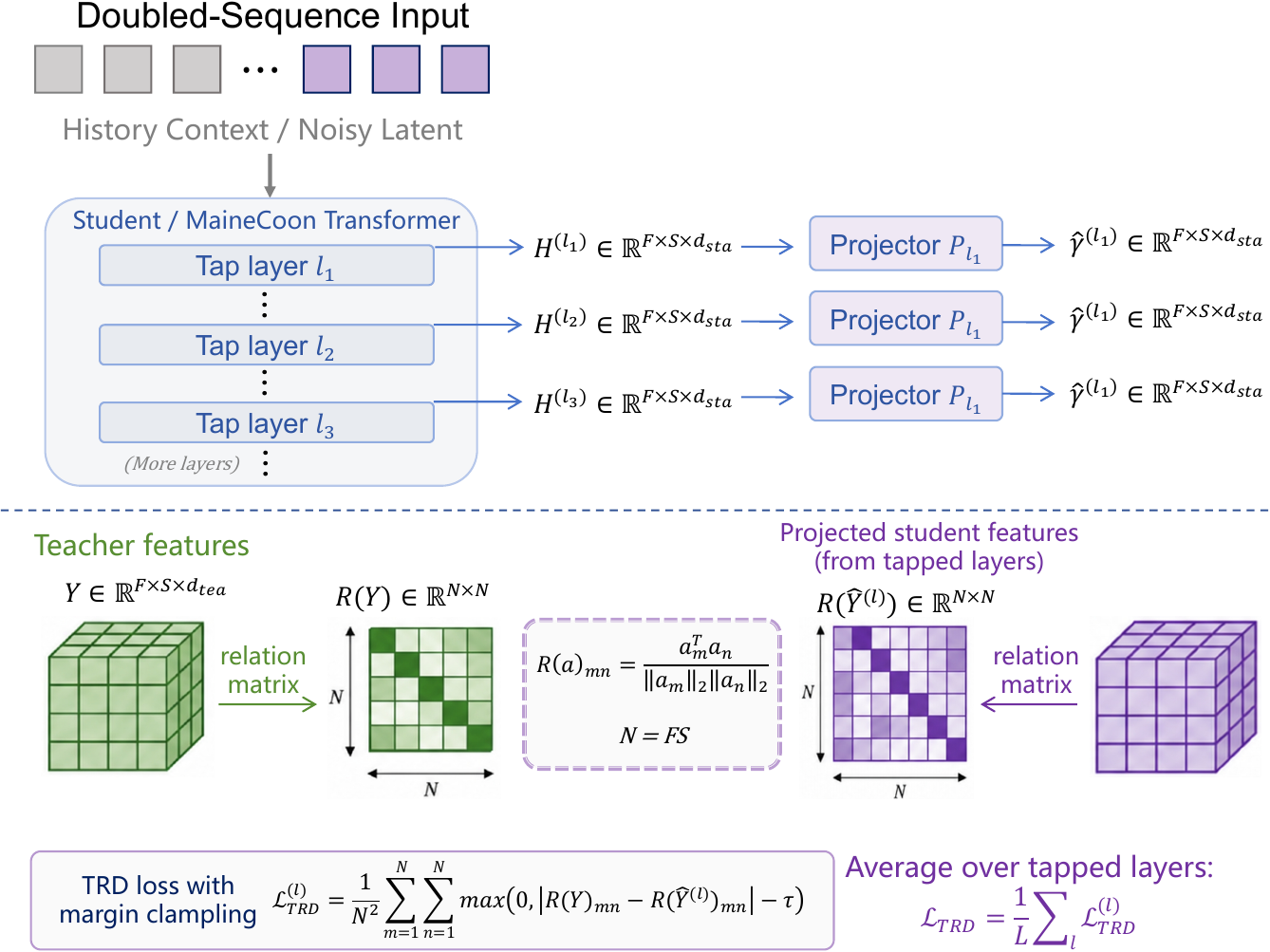}
            \caption{
                Representation alignment accelerates training by matching token
                relation matrices between student visual target features and
                frozen V-JEPA~2 teacher features.
            }
            \label{fig:repa}
        \end{subfigure}
    \end{minipage}

    \vspace{1em}  

    \begin{minipage}[t]{1.0\textwidth}  
        \centering
        \begin{subfigure}[t]{\linewidth}
            \centering
            \includegraphics[width=\linewidth]{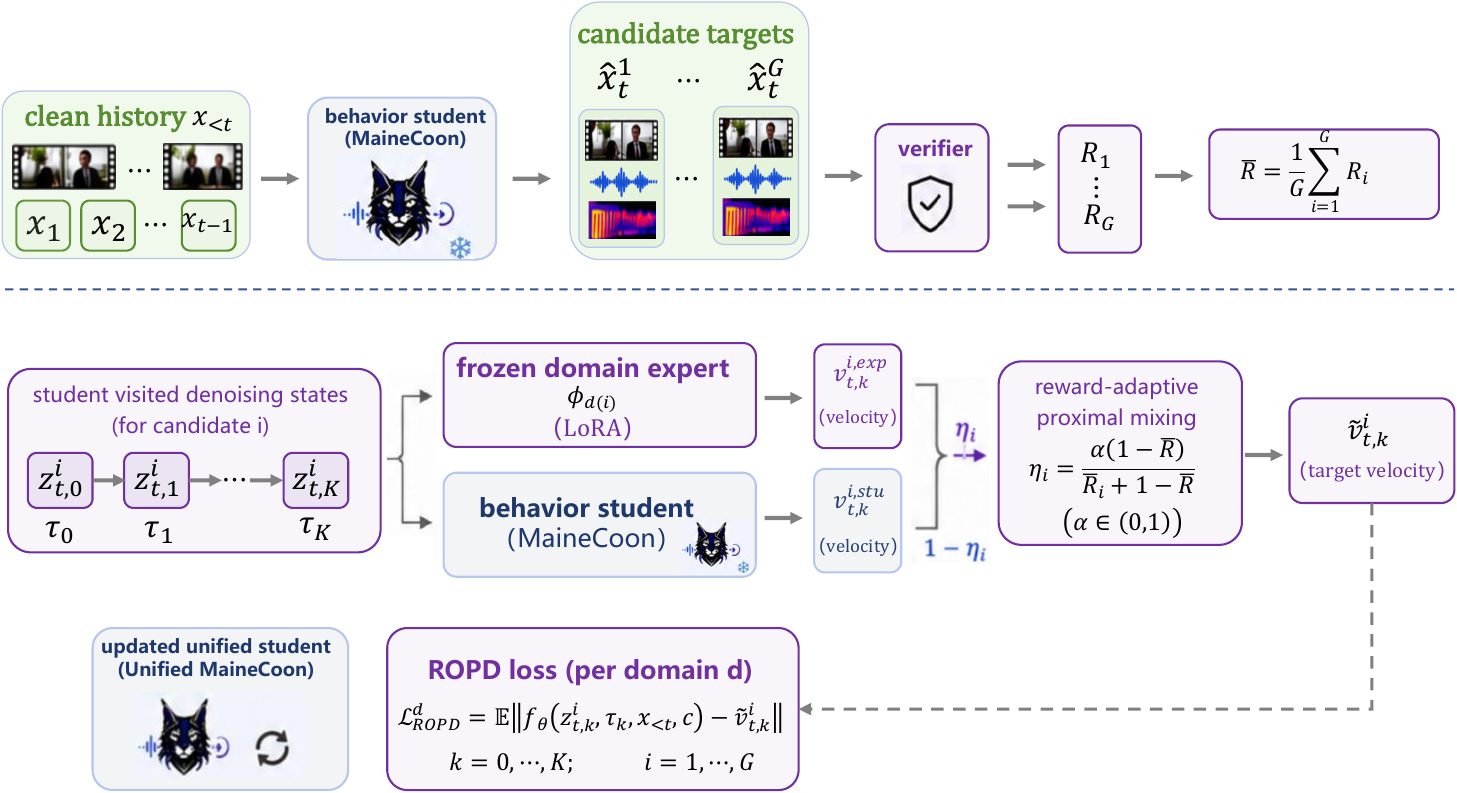}
            \caption{
                Domain-aware preference optimization trains specialized LoRA
                experts and consolidates them into a unified streaming policy
                through reinforced on-policy distillation.
            }
            \label{fig:ropd}
        \end{subfigure}
    \end{minipage}

    \caption{
        \textbf{Training components in MaineCoon.}
        \textbf{(a)} Audio-visual self-resampling exposes the model to degraded
        histories that better match streaming deployment.
        \textbf{(b)} Relational representation alignment provides semantic
        supervision for visual target tokens during the main gradient forward pass.
        \textbf{(c)} Domain-specialized DPO experts are consolidated into one
        unified streaming policy through reward-adaptive policy distillation.
    }
    \label{fig:training_components}
\end{figure}

\paragraph{Limitations of teacher-forced histories}
Training only with clean, real histories creates a context-distribution gap
between training and deployment~\citep{huang2026self, liu2025rolling}. Although the model learns to predict each
chunk from ground-truth past chunks during teacher-forced training, streaming
inference requires it to condition on its own previously generated chunks. As
the rollout grows longer, small errors in identity, motion, audio texture, or
audio-visual synchronization can be repeatedly fed back into the context and
amplified across chunks. Thus, matching the causal attention pattern alone is
not sufficient for stable long-horizon generation; the model must also learn to
recover from imperfect histories similar to those encountered at inference.

\paragraph{Audio-visual self-resampling} Following the motivation of~\citet{guo2025end}, we construct
model-induced degraded histories through self-resampling. As shown in \cref{fig:sr_sampling}, self-resampling supervises each target chunk on a
history that the model generates itself under the deployment-time sampling
procedure. Periodically during training, and under a stop-gradient operator
$\operatorname{sg}[\cdot]$, we run a short autoregressive rollout through the
same causal interface and KV-cache as inference. We draw a history noise level
$\sigma_h\sim p_h$ from a schedule $p_h$ concentrated near $0$ (a clean-biased
level), and produce the past chunks $j=1,\dots,t-1$ one at a time, noising each
chunk and denoising it with a single Euler step conditioned on the chunks
already committed,
\begin{equation}
\begin{aligned}
\mathbf{z}_{j,\sigma_h} &= (1-\sigma_h)\,\mathbf{x}_j+\sigma_h\,\boldsymbol{\epsilon}_j,
\qquad \boldsymbol{\epsilon}_j\sim\mathcal{N}(\mathbf{0},\mathbf{I}),\\
\hat{\mathbf{x}}_j &= \operatorname{sg}\!\big[\,\mathbf{z}_{j,\sigma_h}
-\sigma_h\,f_\theta(\mathbf{z}_{j,\sigma_h},\sigma_h,\hat{\mathbf{x}}_{<j},\mathbf{c})\,\big],
\end{aligned}
\end{equation}
where $\mathbf{z}_{j,\sigma_h}-\sigma_h\,f_\theta(\cdot)$ is the single-step
flow-matching estimate of the clean chunk. The committed chunks
$\hat{\mathbf{x}}_{<t}=(\hat{\mathbf{x}}_1,\dots,\hat{\mathbf{x}}_{t-1})$ form the
self-resampled history. At each training step we choose, per sample, between the
clean and the self-resampled history,
\begin{equation}
\tilde{\mathbf{x}}_{<t}=
\begin{cases}
\mathbf{x}_{<t}, & \text{with probability } 1-\rho,\\[2pt]
\hat{\mathbf{x}}_{<t}, & \text{with probability } \rho,
\end{cases}
\end{equation}
where $\rho\in[0,1]$ is the self-resampling ratio.

We assemble a doubled sequence that places the conditioning history beside the
noised target chunk, and tag the two halves with their noise levels,
\begin{equation}
\mathbf{s}_t=\big[\,\tilde{\mathbf{x}}_{<t}\;\big\Vert\;\mathbf{z}_{t,\tau}\,\big],
\qquad
\boldsymbol{\sigma}(\mathbf{s}_t)=\big[\,\mathbf{0}\;\big\Vert\;\tau\,\big],
\end{equation}
so the history half carries noise level $0$ (a clean tag for the AdaLN
conditioning) while the target half carries the sampled level $\tau$. With this construction, for any history $\mathbf{h}$ we write $f_\theta(\mathbf{z}_{t,\tau},\tau,\mathbf{h},\mathbf{c})$ as shorthand for the target-half velocity that $f_\theta$ predicts when applied to the doubled sequence $[\,\mathbf{h}\,\Vert\,\mathbf{z}_{t,\tau}\,]$, with $\mathbf{h}$ entering only as attention context; this is the meaning of $f_\theta$ in $\mathcal{L}_{\mathrm{AR}}$~(used with $\mathbf{h}=\mathbf{x}_{<t}$), in the rollout above, and in $\mathcal{L}_{\mathrm{SR}}$~(used with $\mathbf{h}=\hat{\mathbf{x}}_{<t}$). Let $c_i$
and $h_i\in\{0,1\}$ denote the chunk index and the half ($0$ for history, $1$
for target) of token $i$. Attention over $\mathbf{s}_t$ follows the block-causal
self-resampling mask
\begin{equation}
M_{ij}=1
\;\Longleftrightarrow\;
\begin{cases}
c_j\le c_i, & h_i=h_j=0,\\
c_j<c_i, & h_i=1,\ h_j=0,\\
c_j=c_i, & h_i=h_j=1,
\end{cases}
\end{equation}
and $M_{ij}=0$ otherwise; in particular the history never attends to the target
($h_i=0,h_j=1$). To reproduce the sliding KV-cache used at inference, which
retains $S$ sink chunks and the $W$ most recent chunks while evicting the rest,
we keep a history key only inside the sink-and-window set
$\mathcal{K}(c_i)=\{c:\,c<S\ \lor\ c\ge c_i-W\}$, giving the sliding-window
doubled-sequence mask
\begin{equation}
M^{\mathrm{sw}}_{ij}=M_{ij}\cdot\big(\mathbf{1}[h_j=1]+\mathbf{1}[h_j=0]\,\mathbf{1}[c_j\in\mathcal{K}(c_i)]\big),
\end{equation}
where $\mathbf{1}[\cdot]$ is the indicator function. We evaluate
$M^{\mathrm{sw}}$ as a FlexAttention block mask rather than a dense $T\times T$
tensor, which keeps the doubled-sequence attention tractable as the rollout
horizon grows.

For both choices of history the loss is taken only on the target half of
$\mathbf{s}_t$; the history half receives no loss, and the self-resampled
rollout carries no gradient. Writing the self-resampled objective explicitly,
\begin{equation}
\mathcal{L}_{\mathrm{SR}}
=\mathbb{E}_{t,\tau,\boldsymbol{\epsilon}}\!\left[
\big\| f_\theta(\mathbf{z}_{t,\tau},\tau,\hat{\mathbf{x}}_{<t},\mathbf{c})-\mathbf{u}_{t,\tau}\big\|_2^2
\right],
\end{equation}
which differs from $\mathcal{L}_{\mathrm{AR}}$ only by conditioning on the
self-resampled history $\hat{\mathbf{x}}_{<t}$ instead of the clean history
$\mathbf{x}_{<t}$. Video and audio share this construction, each contributing a
history half and a noised target half to $\mathbf{s}_t$. Self-resampling
therefore trains the model both to generate high-quality chunks from clean
context and to recover from the degraded context produced by its own rollout.

\paragraph{Curriculum} Because the per-sample history is clean with probability
$1-\rho$ and self-resampled with probability $\rho$, the overall training
objective is the $\rho$-weighted combination
\begin{equation}
\mathcal{L}_{\mathrm{native}}
=(1-\rho)\,\mathcal{L}_{\mathrm{AR}}+\rho\,\mathcal{L}_{\mathrm{SR}}.
\end{equation}
We anneal this objective as a curriculum. Early in training, $\rho$ is small and
the rollout horizon is short, so supervision relies mostly on clean
teacher-forced histories, which establishes local audio-visual quality, speech
clarity, and lip synchronization. As training proceeds, we gradually increase
$\rho$ and extend the rollout horizon, exposing the model to longer and more
imperfect self-resampled histories. This curriculum does not treat
self-resampling as a standalone solution to all long-horizon errors; instead, it
improves the model's robustness to the model-induced context degradation that
naturally appears during streaming inference.

\subsection{Streaming Representation Alignment} \label{subsec:repa}

Training a large audio-visual diffusion transformer from scratch under the
streaming objective of \cref{subsec:native_sr} is slow to acquire semantic
structure: the flow-matching loss supervises low-level reconstruction and
exerts only weak pressure toward semantically organized internal features, so
coherent motion and audio-visual correspondence emerge late in training.
Following REPA~\citep{yu2024repa}, we accelerate training by aligning the
model's intermediate representations to those of a frozen, pretrained
self-supervised video encoder. Because video representations are better
characterized by the relations among tokens than by absolute feature
values, we adopt the relational form of this idea, token relation distillation,
following VideoREPA~\citep{zhang2025videorepa}, with V-JEPA~2~\citep{assran2025vjepa2}
as the teacher. Alignment is applied to the visual stream only; the audio
stream is left unconstrained, as the encoder is visual. An illustration of this alignment process is shown in \cref{fig:repa}.

\paragraph{Teacher representations and grid alignment} We pre-compute teacher
features with a frozen encoder $g$ (V-JEPA~2). For a training clip we sample
frames at the model's frame rate and resize them so that the encoder's patch
grid matches the visual latent grid; the encoder then yields a feature volume
\begin{equation}
\mathbf{Y}=g(\text{clip})\in\mathbb{R}^{F\times S\times d_{\mathrm{tea}}},
\end{equation}
where $F$ is the number of latent frames, $S$ the number of spatial tokens per
frame, and $d_{\mathrm{tea}}$ the teacher channel dimension, with teacher tokens
placed in one-to-one correspondence with the visual latent tokens. When the
encoder's temporal resolution differs from the latent frame count, we reconcile
it to $F$ by trilinear interpolation of the teacher volume.

\paragraph{Token relation distillation} At a set $\mathcal{A}$ of intermediate
transformer layers, inspired by~\citep{zhang2025videorepa}, we tap the hidden states of the noisy visual target tokens
and map them into the teacher space with a per-layer convolutional projector
$P_\ell$ that preserves the spatial grid,
$\hat{\mathbf{Y}}^{(\ell)}=P_\ell\!\big(\mathbf{H}^{(\ell)}\big)\in\mathbb{R}^{F\times S\times d_{\mathrm{tea}}}$.
Treating a feature volume as a set of $N=F\,S$ token vectors
$\mathbf{a}\in\mathbb{R}^{N\times d}$, we summarize its structure by the cosine
relation matrix
\begin{equation}
R(\mathbf{a})_{mn}=\frac{\mathbf{a}_m^{\top}\mathbf{a}_n}{\|\mathbf{a}_m\|_2\,\|\mathbf{a}_n\|_2},
\qquad m,n=1,\dots,N,
\end{equation}
which captures the pairwise similarity of all tokens independently of the basis
in which they live. Token relation distillation matches the student and teacher
relation matrices through a margin-clamped error,
\begin{equation}
\ell_{\mathrm{TRD}}\!\big(\hat{\mathbf{Y}},\mathbf{Y}\big)
=\frac{1}{N^2}\sum_{m,n=1}^{N}
\operatorname{ReLU}\!\Big(\big|R(\hat{\mathbf{Y}})_{mn}-R(\mathbf{Y})_{mn}\big|-\gamma\Big),
\end{equation}
where $\gamma\ge 0$ is a hinge margin that ignores small relational
discrepancies. Aligning relations rather than raw features lets the generator
keep its own representation basis as long as the pairwise token structure
agrees, which is more stable than direct feature regression for video. The same loss is applied at several mid-to-late transformer layers; each tapped layer has its own projector and is aligned independently to the same teacher features, and the per-layer terms are averaged into the alignment loss $\mathcal{L}_{\mathrm{align}}$. Aligning several layers rather than a single one keeps every block on an alignment-gradient path.

\paragraph{Integration with native streaming training} Representation alignment
is an auxiliary objective added to the native streaming training of
\cref{subsec:native_sr}; the full training objective is
\begin{equation}
\mathcal{L}=\mathcal{L}_{\mathrm{native}}+\lambda_{\mathrm{align}}\,\mathcal{L}_{\mathrm{align}},
\end{equation}
with alignment weight $\lambda_{\mathrm{align}}$. The teacher $g$ is frozen and
its features are pre-computed, so alignment adds no teacher forward pass at
training time. The loss is applied only to the visual target half of the doubled
sequence and only on the main gradient forward pass; it is disabled during the
no-gradient self-resampling rollout. Because the teacher supplies semantically
organized features from the start, the model reaches coherent motion and
audio-visual correspondence in far fewer steps, accelerating training while
also improving generation quality.

\subsection{Posttraining: Multi-domain DPO and ROPD} 
\label{subsec:dpo_ropd} 

The native streaming objective learns a unified causal audio-visual generator, but the quality criteria of social video vary substantially across content domains. For example, far-shot videos require stable full-body structure, multi-person dialogue requires consistent speaker identity and turn-taking, while dance and high-motion videos emphasize large, temporally coherent body movement. Directly optimizing all these objectives in one model can introduce conflicting preference signals. Inspired by the success of on-policy distillation in both LLMs~\citep{agarwal2024policy,song2026survey,fu2026revisiting,deepseek2026deepseek} and diffusion models~\citep{li2026diffusionopd}, which provides denser learning signals than single-stage preference learning~\citep{liu2026alignment,zheng2025diffusionnft,wallace2024diffusion,liu2026flow}, we therefore adopt a specialize-then-consolidate strategy. As shown in~\cref{fig:ropd}, we first construct domain-specific preference pairs and train a bank of lightweight DPO experts. We then consolidate these experts into one deployable streaming policy through reinforced on-policy distillation.

\paragraph{Domain preference pairs}
We consider at least five challenging domains:
\textit{far shot}, \textit{multi-person dialogue}, \textit{motion},
\textit{animation}, and \textit{dance}. For each domain $d\in\mathcal{D}$, we
select high-quality real videos using domain-specific quality filters and treat
them as preferred samples $\mathbf{x}^{+}$. Given the structured prompt
$\mathbf{c}$ extracted from each preferred video, the current generator produces
a dispreferred sample
\begin{equation}
\mathbf{x}^{-}\sim
p_{\theta_{\mathrm{gen}}}(\mathbf{x}\mid\mathbf{c}).
\end{equation}
The preferred and dispreferred videos share the same prompt and form a
preference pair $(\mathbf{c},\mathbf{x}^{+},\mathbf{x}^{-})$. We periodically
refresh the generated negatives using the latest domain model, so that the
preference data continue to reflect its current failure modes.

\paragraph{Domain-specialized DPO experts} Starting from the native streaming checkpoint $\theta_0$, we train one LoRA expert for each domain, 
\begin{equation} \phi_d = \theta_0+\Delta_d, \qquad d\in\mathcal{D}, 
\end{equation} 
where $\Delta_d$ denotes the trainable domain adapter. For a sampled chunk $t$, the preferred and dispreferred samples are perturbed using the same noise level $\tau$ and Gaussian noise $\boldsymbol{\epsilon}$, 
\begin{equation} \mathbf{z}^{y}_{t,\tau} = (1-\tau)\,\mathbf{x}^{y}_t + \tau\,\boldsymbol{\epsilon}, \qquad \mathbf{u}^{y}_{t,\tau} = \boldsymbol{\epsilon}-\mathbf{x}^{y}_t, \qquad y\in\{+,-\}, \end{equation}
where 
\begin{equation} \boldsymbol{\epsilon} \sim \mathcal{N}(\mathbf{0},\mathbf{I}). 
\end{equation} 

The two candidates are represented using the same doubled-sequence interface as native streaming training, \begin{equation} 
\mathbf{s}^{+}_t = \big[\, \mathbf{x}^{+}_{<t} \;\big\Vert\; \mathbf{z}^{+}_{t,\tau} \,\big], \qquad \mathbf{s}^{-}_t = \big[\, \mathbf{x}^{-}_{<t} \;\big\Vert\; \mathbf{z}^{-}_{t,\tau} \,\big]. 
\end{equation} 

They share the prompt, chunk index, noise level, and sliding block-causal mask $M^{\mathrm{sw}}$, while each retains its own clean trajectory history. As in native streaming training, the loss is evaluated only on the target half. For $y\in\{+,-\}$, we define the target-half flow-matching error as 

\begin{equation} \ell_{\theta}^{y} = \left\| f_\theta \left( \mathbf{z}^{y}_{t,\tau}, \tau, \mathbf{x}^{y}_{<t}, \mathbf{c} \right) - \mathbf{u}^{y}_{t,\tau} \right\|_2^2. 
\end{equation}

Using the frozen native streaming checkpoint $\theta_0$ as the reference model, the domain preference objective is \begin{equation} \mathcal{L}_{\mathrm{DPO}}^{d} = -\mathbb{E}\!\left[ \log\sigma\!\left( \beta_d \left[ \left( \ell_{\phi_d}^{-} - \ell_{\phi_d}^{+} \right) - \left( \ell_{\theta_0}^{-} - \ell_{\theta_0}^{+} \right) \right] \right) \right], 
\end{equation} 
where $\beta_d$ controls the preference strength. We additionally retain a small reconstruction loss on the preferred samples, 
\begin{equation} \mathcal{L}_{\mathrm{expert}}^{d} = \mathcal{L}_{\mathrm{DPO}}^{d} + \lambda_{\mathrm{win}}\, \mathbb{E}\!\left[ \ell_{\phi_d}^{+} \right]. 
\end{equation} 

The resulting adapters form a frozen domain-expert bank $\{\phi_d\}_{d\in\mathcal{D}}$.

\paragraph{Reinforced on-policy distillation}
Directly averaging the domain adapters may cause interference, while routing
multiple experts during inference increases deployment complexity. We
therefore consolidate their capabilities into a single student initialized
from the native streaming checkpoint, $\theta\leftarrow\theta_0$.

For a training sample from domain $d$, we retain the clean ground-truth history
$\mathbf{x}_{<t}$ and use the behavior student to generate a group of $G$
candidate target chunks,
\begin{equation}
\hat{\mathbf{x}}^{\,i}_t
\sim
p_{\theta_{\mathrm{old}}}
\left(
\mathbf{x}_t
\mid
\mathbf{x}_{<t},
\mathbf{c}
\right),
\qquad
i=1,\dots,G.
\end{equation}
Each candidate is produced through a student-generated denoising trajectory
$\{\mathbf{z}^{i}_{t,k}\}_{k=0}^{K}$. At denoising step $k$, the corresponding
doubled sequence is
\begin{equation}
\mathbf{s}^{i}_{t,k}
=
\big[\,
\mathbf{x}_{<t}
\;\big\Vert\;
\mathbf{z}^{i}_{t,k}
\,\big],
\end{equation}
which follows the same sliding block-causal mask $M^{\mathrm{sw}}$ used in
native streaming training. The frozen domain expert is evaluated only on these
student-visited states and does not generate a separate trajectory.

A domain-specific verifier assigns a binary outcome $R_i\in\{0,1\}$ to each
candidate, and the group success rate is
\begin{equation}
\bar{R}
=
\frac{1}{G}
\sum_{i=1}^{G}R_i.
\end{equation}

We propose reinforced on-policy distillation (ROPD) to combine the
candidate-level outcome with a proximal domain-expert prior. Its unified
transition-level reward is
\begin{equation}
r^{\mathrm{ROPD}}_{i,t,k}
=
\log\!\left(
R_i
+
(1-\bar{R})
\left(
\alpha\,\rho_{i,t,k}
+
1-\alpha
\right)
\right),
\label{eq:ropd_reward}
\end{equation}
where $\rho_{i,t,k}$ denotes the domain-expert-to-student transition ratio and
$\alpha\in(0,1)$ controls the maximum contribution of the expert.

To expose the proximal structure of this reward, we define
\begin{equation}
Z_i
=
R_i+1-\bar{R},
\qquad
\eta_i
=
\frac{
\alpha(1-\bar{R})
}{
R_i+1-\bar{R}
}.
\end{equation}
The reward in \cref{eq:ropd_reward} can then be written as
\begin{equation}
r^{\mathrm{ROPD}}_{i,t,k}
=
\log Z_i
+
\log\!\left(
(1-\eta_i)
+
\eta_i\,\rho_{i,t,k}
\right).
\end{equation}
Therefore, $\eta_i$ is the posterior weight assigned to the corresponding
domain expert. Unlike a manually selected distillation coefficient, it is
determined jointly by the candidate outcome and the empirical success rate of
the current student.

Inspired by DiffusionOPD~\citep{li2026diffusionopd}, we optimize this ROPD
target directly on student-visited denoising states rather than converting it
into a PPO-style policy-gradient objective. For the deterministic flow model
used by MaineCoon, we realize the induced proximal target in velocity space,
\begin{equation}
\begin{aligned}
\widetilde{\mathbf{v}}^{i}_{t,k}
={}&
(1-\eta_i)\,
\operatorname{sg}\!\left[
f_{\theta_{\mathrm{old}}}
\left(
\mathbf{z}^{i}_{t,k},
\tau_k,
\mathbf{x}_{<t},
\mathbf{c}
\right)
\right]\\
&+
\eta_i\,
\operatorname{sg}\!\left[
f_{\phi_d}
\left(
\mathbf{z}^{i}_{t,k},
\tau_k,
\mathbf{x}_{<t},
\mathbf{c}
\right)
\right].
\end{aligned}
\end{equation}
The domain-specific ROPD objective is
\begin{equation}
\mathcal{L}_{\mathrm{ROPD}}^{d}
=
\mathbb{E}_{i,t,k}\!\left[
\left\|
f_\theta
\left(
\mathbf{z}^{i}_{t,k},
\tau_k,
\mathbf{x}_{<t},
\mathbf{c}
\right)
-
\widetilde{\mathbf{v}}^{i}_{t,k}
\right\|_2^2
\right].
\end{equation}
This pathwise objective preserves the reward-adaptive expert intervention of
ROPD while avoiding stochastic transition ratios and PPO-style optimization.

If all candidates fail, then $R_i=\bar{R}=0$ and $\eta_i=\alpha$, so every
candidate is trained toward a proximal mixture of the behavior student and the
domain expert. If all candidates succeed, then $\bar{R}=1$ and $\eta_i=0$, so
the expert provides no additional supervision. For a mixed group, failed
candidates receive the maximum expert weight $\alpha$, whereas successful
candidates receive a smaller expert weight that decreases as the group success
rate increases.

\paragraph{Multi-domain consolidation} During each training round, we visit the five domains in a balanced order and use the corresponding frozen LoRA as the teacher. The domain losses are averaged before updating the unified student, \begin{equation} \mathcal{L}_{\mathrm{round}} = \frac{1}{|\mathcal{D}|} \sum_{d\in\mathcal{D}} \mathcal{L}_{\mathrm{ROPD}}^{d}. \end{equation} We additionally retain the native streaming objective as a real-data anchor, \begin{equation} \mathcal{L}_{\mathrm{post}} = \mathcal{L}_{\mathrm{round}} + \lambda_{\mathrm{native}}\, \mathcal{L}_{\mathrm{native}}. \end{equation} After training, all domain adapters and verifiers are removed, and inference uses only one unified MaineCoon policy with the original streaming architecture and deterministic few-step sampler.

\paragraph{Step Distillation} Additionally, we combine DMD~\citep{yin2024improved} and its two variants~\citep{shao2025magicdistillation,bai2026optimizing} to achieve nearly-lossless four-step distilled version of MaineCoon base model. Our method can prevent unstable gradient signals in the early stage of step distillation by softening the predictions of the real teacher during this phase, which is highly desired for distilling large-scale video generation models. The step-distilled weight bias can be also loaded to other posttrained Maine Coon ckpts. When combined with the agentic streaming inference strategy in \cref{sec:inference}, which governs chunk
commitment, KV-cache reuse, and long-context rollout, native streaming training
forms a train-inference matched pipeline for stable long-horizon real-time audio-visual
generation without relying on non-causal teachers.

\subsection{Training Infrastructure} \label{subsec:training_infra}

\paragraph{Parallelism} We train the full 22B audio-visual transformer with a
combination of fully sharded data parallelism and sequence parallelism across
64 H100 GPUs. Parameters, gradients, and optimizer states are
sharded with FSDP2 (per-parameter \texttt{DTensor} sharding), taking each
audio-visual transformer block as the wrapping unit. On top of this,
Ulysses-style sequence parallelism (degree $\mathrm{SP}=4$) splits each long
video token sequence across four ranks; these four-way groups are kept
within a single node so their all-to-all exchanges stay on NVLink. Together this
bounds per-rank activation memory and lets the long chunked rollouts of
\cref{subsec:native_sr} fit on device.

\paragraph{Optimization} We initialize from the LTX-2.3 checkpoint~\citep{hacohen2026ltx}, a 22B-parameter open-source audio-visual diffusion model, and then optimize with 8-bit AdamW~\citep{dettmers20218,loshchilov2018decoupled,kingma2014adam},
$(\beta_1,\beta_2)=(0.9,0.999)$, weight decay $0.01$, and gradient-norm clipping
at $1.0$. The learning rate is held constant at $1\times10^{-5}$ after a short
linear warmup. Because each training example is a long audio-visual rollout, we
use a per-device micro-batch of one and accumulate gradients over three steps;
an exponential moving average of the weights (decay $0.999$) is maintained and
used as the final model. In total, our training consumes only 10k-scale GPU hours and uses only 1M-scale extreme high-quality long videos, covering diverse social video domains. 

\paragraph{Training Strategies} Several systems-level choices make
22B-scale streaming training tractable on H100. Computation runs in bf16 mixed
precision, and the FSDP2 shards offload their fp32 master parameters to pinned
CPU memory, freeing device memory for the long doubled sequences. Activation
memory is controlled by gradient checkpointing applied inline within each
transformer block, rather than wrapping the block a second time, which would
needlessly multiply the backward recompute. The optimizer states are kept in
8-bit, and the text encoder, needed only to embed validation prompts, is loaded
in 8-bit. The video and audio latents, text embeddings, and the V-JEPA teacher
features of \cref{subsec:repa} are all precomputed offline, so the training loop
runs no VAE, text-encoder, or teacher forward pass and instead streams these
tensors from disk using pinned-memory transfers and persistent dataloader workers. Finally,
training clips are bucketed by their latent-frame count and all ranks draw from the same bucket at each step, so every micro-batch holds equal-length sequences across all ranks; this keeps the FSDP all-gather sizes uniform and wastes no computation on padding.

\section{Agentic Streaming Inference}
\label{sec:inference}

A real-time social experience is not a finished clip but an open-ended exchange: the viewer talks, types, or simply watches, and the stream must answer at once and never end. The frozen MaineCoon generator emits audio-visual chunks at real-time speed, but on its own it cannot plan what comes next, retain what it has already shown, or react to the viewer while a chunk is in flight. We therefore introduce a training-free agentic streaming inference framework that stands between the viewer and the generator and turns it into a minute-scale, interactive stream, to our knowledge the first such framework for real-time audio-visual autoregressive generation.

\begin{figure}[tp]
    \centering
    \IfFileExists{Assets/Figures/agentic_inference.pdf}{%
      \includegraphics[width=\linewidth]{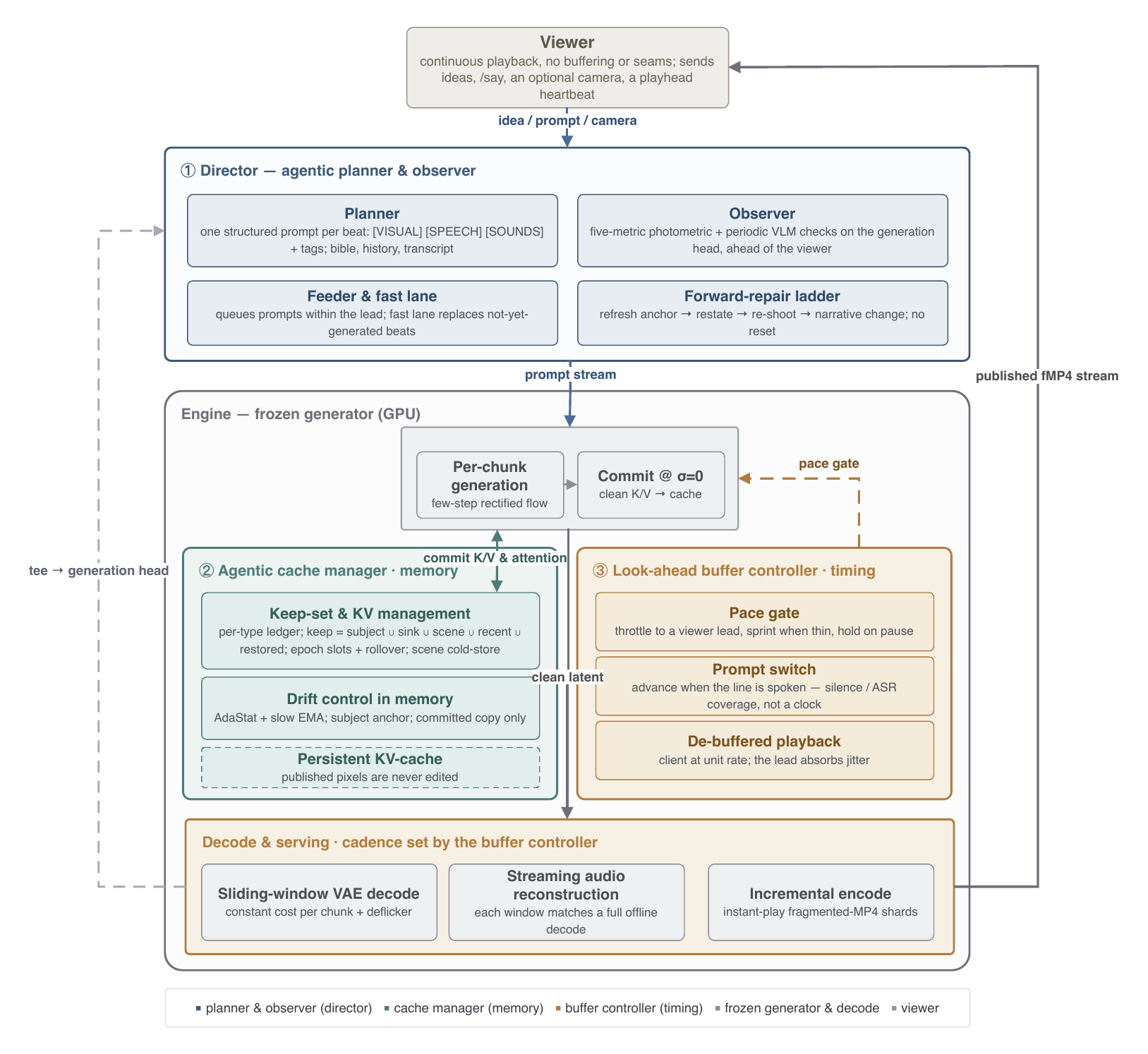}}{%
      \fbox{\parbox[c][2.4cm][c]{0.985\linewidth}{\centering\small\itshape Figure pending: export \texttt{agentic_inference.pdf} from the agentic-inference tab of \texttt{Assets/Figures/data-pipeline.drawio}.}}}
    \caption{\textbf{Agentic streaming inference.} A training-free framework wraps the frozen MaineCoon generator, shown in grey, and adds three agentic controllers, each drawn in its own color. The \textcolor{blockPlan}{agentic planner and observer}, the director, writes one structured prompt per beat, watches the generation head ahead of the viewer, and repairs degradation forward through a graded ladder without ever resetting the stream. The \textcolor{blockCache}{agentic cache manager} governs the engine's memory: each audio-visual chunk is committed once to a never-cleared KV-cache, where a bounded keep-set holds the context inside the trained regime and in-memory drift anchors act only on the committed copy, so the published pixels remain the generator's native output. The \textcolor{blockBuf}{agentic look-ahead buffer controller} governs timing: a pace gate holds the lead of generated-but-unwatched content within bounds while a sliding-window decoder serves the stream continuously to the viewer, who can steer the show in real time.}
    \label{fig:agentic_system}
\end{figure}

Bridging the viewer to the generator takes three jobs, and the framework meets each with one agentic controller that governs a single axis without retraining. The first job is to author the show: on the director side, an agentic planner and observer writes the prompt stream that drives endless generation, continues the story beat by beat, and folds viewer interaction into the next beat. The second is to remember: on the engine side, an agentic cache manager governs the KV-cache, carrying continuity natively across an unbounded stream and correcting the slow appearance drift that any long stream accumulates. The third is to keep time: because the generator runs faster than playback, a lead of generated-but-unwatched content builds up ahead of the viewer, a problem still largely unexamined for interactive generation, and an agentic look-ahead buffer controller turns this lead into a resource for picture quality and early monitoring while holding interaction responsive. The architecture of our agentic streaming system is depicted in~\cref{fig:agentic_system}.

\subsection{Agentic Planner and Observer}
\label{subsec:planner}

The prompt stream that drives the whole system is produced by an agentic planner and observer, a locally deployed Gemma 4 26B mixture-of-experts model~\citep{gemma4} that runs the session as an agent and is the cognitive core of the framework, carrying out all of the system's reasoning while the engine-side controllers only execute fixed policies. It acts in two roles over a single prompt channel: as planner it authors the prompt stream that drives generation, and as observer it watches the generated stream and repairs degradation forward. It never starts, stops, or steps the generator, which continues at its fixed cadence, so content and memory stay on the agent side and timing on the engine side, a separation that lets the system respond to the viewer without ever interrupting the stream.

\paragraph{Planner} At each beat the planner writes one self-contained prompt, a visual description that restates a fixed character specification, a single line of speech, and the ambient sound, that advances the story while reusing the previous shot's framing; it maintains a bounded planning history and a spoken-line transcript so that the show never ends and never repeats itself.

\paragraph{Feeder and fast lane} A feeder queues these prompts asynchronously and keeps the queue within the look-ahead lead, exposing a fast lane that replaces only the not-yet-generated beats, so that whatever the agent decides next reaches the engine without interrupting the chunk in flight.

\paragraph{Observer} The same agent also observes the stream for degradation as it is generated. Because generation leads playback, this observation runs on the generation head, on content the viewer has not yet seen, so a correction is in flight before the defect would reach the screen. A cheap five-metric photometric drift score flags appearance drift, and a periodic vision-language check confirms semantic defects such as identity or wardrobe drift. What it detects, it repairs forward rather than by resetting the stream.

\paragraph{Forward-repair ladder} Detected degradation is handled by a graded and entirely forward repair ladder: refresh the subject anchor, restate the canonical appearance in the next prompt, re-shoot the beat, and, if degradation persists, drive a narrative change that walks the story out of the degraded state so that fresh content dilutes the stale context. The ladder escalates only as far as needed and contains no hard reset, because a reset is itself the most visible discontinuity a viewer can perceive; repairing forward instead removes the tearing that periodic re-anchoring would cause.

\paragraph{Viewer interaction} The same prompt channel carries viewer interaction. A typed instruction, and optionally a camera feed, is folded into the next plan and lands at the next natural switch point without interrupting the stream. Every auxiliary path is built to degrade gracefully, so that if segmentation, the vision-language check, or the planner fails, the system falls back to a coarser signal or a safe continuation and no observer-side failure can ever freeze the stream.

\subsection{Agentic Cache Manager}
\label{subsec:cache}

The agentic cache manager governs everything the model remembers. At inference each chunk is produced by the same few-step rectified-flow procedure used in training and is then committed to memory: a final pass at zero noise level writes the chunk's clean keys and values into a KV-cache that persists across the entire stream, after which the emitted latents are frozen and never revised. This commit is the single point through which information enters the model's memory, which makes it the natural place to govern two things over one persistent cache: how much the model remembers, which must stay bounded and inside the trained regime, and how faithfully it remembers, which must resist drift.

\paragraph{One persistent cache} A long stream could be produced in two obvious ways, both of which we reject. Regenerating each chunk from the growing prefix makes the per-chunk cost grow without bound; stitching independently generated segments and periodically re-anchoring them introduces visible seams and lets identity, color, and audio timbre jump at every segment boundary. Instead we maintain one continuous stream over a single KV-cache that is never cleared, so inter-chunk continuity is carried natively by attention and there are no segment boundaries to bridge.

\paragraph{Bounded keep-set} A persistent cache cannot be allowed to grow without limit, because the generator was trained on bounded clips with a bounded range of rotary positions, so both compute and the position encoding break once the cache accumulates past the training horizon. The manager therefore keeps the cache small and inside the trained regime while preserving the content on which long-horizon consistency depends. After each commit it recomputes a small keep-set and evicts everything else. The generator uses four attention types, namely visual and audio self-attention and the two cross-modal audio-visual paths, whose caches grow at different rates because audio-free chunks append only to the visual self-attention path; a per-attention-type ledger tracks these counts separately so that eviction never misaligns the modalities. The retained set is non-contiguous,
\begin{equation}
\mathcal{K}_t=
\underbrace{\mathcal{G}}_{\text{subject anchors}}\ \cup\
\underbrace{\mathcal{S}}_{\text{scene sink}}\ \cup\
\underbrace{\mathcal{A}}_{\text{scene anchors}}\ \cup\
\underbrace{\mathcal{R}_t}_{\text{recent chunks}}\ \cup\
\underbrace{\mathcal{O}_t}_{\text{restored on return}},
\end{equation}
combining a persistent attention sink from the scene's establishing chunk, a few scene anchors, the most recent chunks, the subject anchors introduced below, and any chunks restored when a scene returns; every other token is dropped, and each surviving token retains its original positional encoding so that it still maps to its true moment in time. A fixed budget of a few recent chunks caps the cache, which makes throughput constant regardless of how long the stream has run.

\paragraph{Bounded positions} Two further mechanisms keep positions inside the trained range. Positions are assigned within bounded epoch slots that always fall within the training horizon; when the slots of an epoch are exhausted, the cache is rebuilt from the retained clean latents at fresh slots, which restores the position encoding to the training distribution while preserving content, because what is replayed is the stream's own committed history. Scene changes are handled through memory rather than through re-anchoring: when the prompt declares that a scene leaves, its establishing chunks are moved to host storage and the cache is cleared, and when the scene returns they are restored at the front with their original positions. Scene boundaries are taken only from the prompt stream and are never inferred from generated frames, which prevents the manager from depending on the very output it is meant to stabilize. Together, these mechanisms let a generator trained on roughly twenty-second clips run as one continuous stream for forty-five minutes with no measurable degradation.

\paragraph{Drift control} The same cache is where drift is corrected. Even over a single persistent cache, a streamed subject slowly departs from its intended appearance, color, and audio timbre; we call this accumulated departure drift. Our governing principle for controlling it is that the output is never touched: a correction is applied only to the copy of each chunk that is committed into the cache, so the feedback loop that would amplify drift is broken inside the model's context, while the published video remains the generator's native output. Two complementary anchors apply this principle on the committed copy, one stabilizing global appearance and one stabilizing identity.

\paragraph{Statistical anchor} The first correction is a statistical context anchor. At the commit pass, and before writing the chunk to the cache, we match the per-channel statistics of its clean latent to a running reference,
\begin{equation}
\operatorname{AdaStat}(\mathbf{x};\boldsymbol{\mu}^{\star},\boldsymbol{\sigma}^{\star})
=\boldsymbol{\sigma}^{\star}\odot
\frac{\mathbf{x}-\boldsymbol{\mu}(\mathbf{x})}{\boldsymbol{\sigma}(\mathbf{x})}
+\boldsymbol{\mu}^{\star},
\end{equation}
applied per channel and, for audio, at reduced strength. The reference $(\boldsymbol{\mu}^{\star},\boldsymbol{\sigma}^{\star})$ is initialized from the scene's opening and updated by a slow exponential moving average over low-drift chunks only, so that it tracks legitimate scene evolution rather than opposing it. Because the match acts on the committed copy and not on the emitted latent, it suppresses drift without the chunk-boundary luminance pulses and the motion damping that an output-side normalization would introduce.

\paragraph{Subject anchor} The statistical anchor stabilizes global appearance but not identity. For identity we add a semantic subject anchor: a reference latent block of the main subject is warmed into the cache but never decoded to the output, so every later chunk can attend to a stable record of who is on screen. The block is selected without any geometric prior; an open-vocabulary segmenter scores the subject region named by the planner's textual subject description on a periodic snapshot, and a fixed number of the highest-scoring latent tokens is harvested directly from the stream's own clean latents. Harvesting from the stream keeps the anchor training-free, and fixing the token count avoids the recompilation that a variable shape would trigger. The anchor changes the nature of drift: in same-seed comparisons an unanchored stream degrades monotonically, whereas an anchored stream that has drifted is pulled back toward the reference, so drift becomes recoverable rather than irreversible. We keep the anchor set small, a permanent opening anchor and a single rolling refresh, because anchors whose weight exceeds the recent-context window pull the subject backward against its own motion and create tearing.

\subsection{Agentic Look-Ahead Buffer Controller}
\label{subsec:buffer}

\begin{figure}[t]
    \centering
    \IfFileExists{Assets/Figures/data-pipeline-fig-lookahead.pdf}{%
      \includegraphics[width=\linewidth]{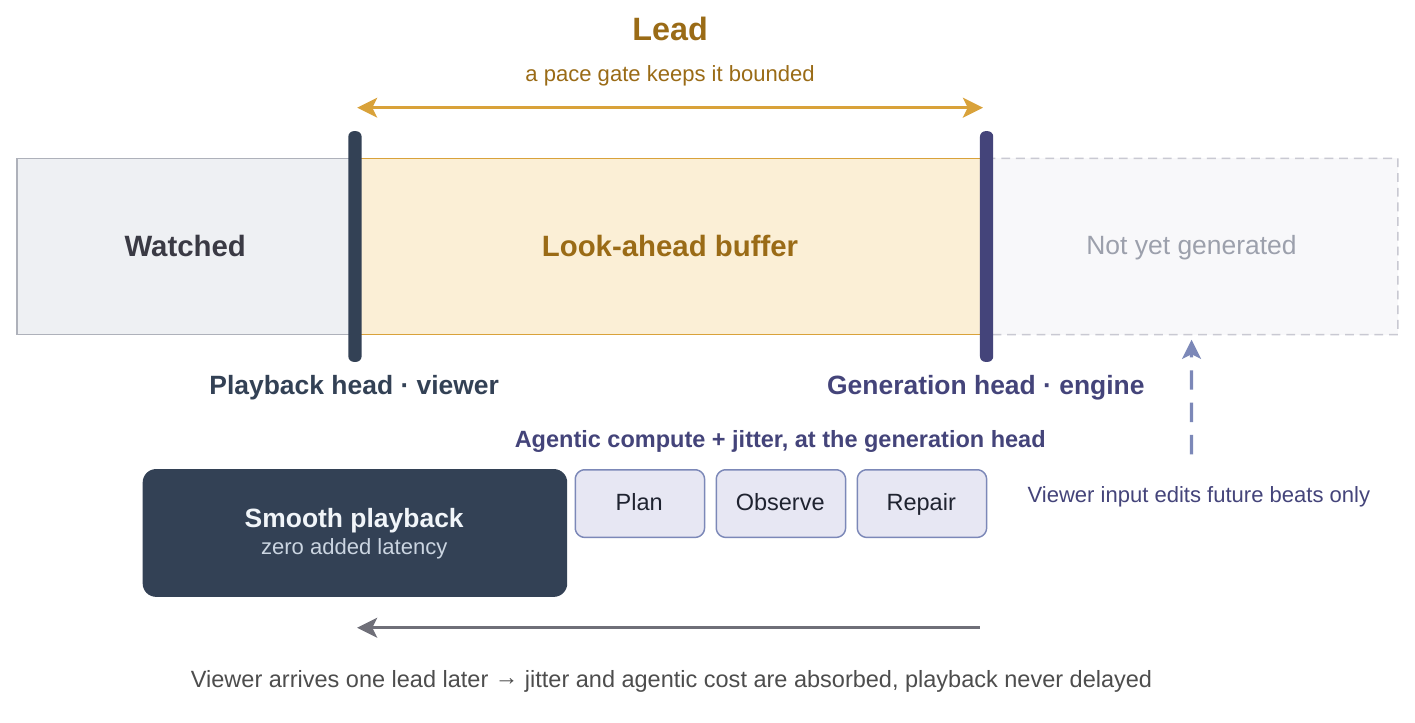}}{%
      \fbox{\parbox[c][2.2cm][c]{0.985\linewidth}{\centering\small\itshape Figure pending: export \texttt{data-pipeline-fig-lookahead.pdf} from the look-ahead-buffer tab of \texttt{Assets/Figures/data-pipeline.drawio}.}}}
    \caption{\textbf{The look-ahead buffer.} Because MaineCoon generates faster than playback, a lead of generated-but-unwatched content opens between the playback head, where the viewer is, and the generation head, where the engine is; the look-ahead buffer controller's pace gate holds this buffer within bounds. All agentic compute, namely planning, observation, and forward repair, is spent at the generation head, inside the buffer and ahead of the viewer, alongside any generation jitter. The viewer reaches that content only one lead later, so the cost is absorbed by the buffer and never delays playback, which stays continuous, unit-rate, and seam-free with zero added latency. A user instruction edits only the not-yet-generated beats and takes effect at the next switch point.}
    \label{fig:lookahead}
\end{figure}

The third controller, the agentic look-ahead buffer controller, governs the pace of generation relative to playback. A streaming generator that is merely fast enough for real time and one that runs comfortably faster than real time behave differently once deployed. On a single GPU MaineCoon samples at about 32 frames per second, above the 25 frames per second at which the stream is played back, as reported in~\cref{tab:latency_master}, so generation and playback do not advance in lockstep: a lead of already-generated but not-yet-watched content accumulates ahead of the viewer, as~\cref{fig:lookahead} illustrates. We observe that this lead, which we call the look-ahead buffer, is itself a design surface, and that managing it is a distinct and largely unexamined problem for interactive streaming generation. Because the buffer both buys time and commits the future, the controller holds it in a sweet spot with three mechanisms, on top of a decode path fast enough to sustain the lead in real time.

\paragraph{Asset and liability} As an asset, the look-ahead buffer is time bought in advance: it absorbs generation jitter so playback never stalls, and it is a window in which the system can act before the viewer is affected, so the observation and forward repair run on the generation head and correct content the viewer has not yet reached. As a liability, the buffer is committed future: a viewer instruction cannot alter frames that are already generated and queued, so every second of lead is a second added to the visible response time. A larger buffer thus buys smoothness and look-ahead at the cost of responsiveness, and the central question is how to hold the lead long enough to protect the viewing experience yet short enough to keep interaction prompt. The controller answers this with three mechanisms.

\paragraph{Pace gate} A pace gate regulates the lead from the viewer's playhead: it throttles generation when the lead exceeds a target window and sprints, generating faster than real time, when the buffer runs thin, so the lead neither stalls nor runs away and the worst-case interaction delay stays bounded; when the viewer pauses, the engine holds the lead rather than racing ahead.

\paragraph{Prompt switch} Prompts advance on a spoken-signal switch: the engine moves to the next queued prompt only when the current line has actually been spoken, judged from the decoded audio by trailing silence after the expected duration or by the transcribed coverage of the scripted line rather than by a clock, so that speech is never cut short or looped; viewer steering enters on the same fast lane and replaces only the not-yet-generated beats, so a command takes effect at the generation head within roughly the planning latency plus the remainder of the current utterance and reaches the viewer one lead later, not after the whole queue drains.

\paragraph{De-buffered playback} Playback is de-buffered: the client keeps no prebuffer gate and plays at unit rate, showing each shard as soon as it arrives, so that the engine-side lead, not a client-side buffer, absorbs jitter and keeps playback seam-free. Together these three mechanisms turn the look-ahead buffer from an incidental by-product of fast generation into a controlled resource that improves the viewing experience while keeping the system responsive.

\paragraph{Real-time decoding} Underneath the control logic, sustaining the lead in real time on a single GPU requires the decoder and the generator to keep pace with the fixed chunk cadence. We decode each chunk with a sliding-window VAE whose cost is constant in stream length rather than growing with the prefix, compile the transformer block by block to cut per-chunk denoising latency, and run decoding, video encoding, and audio reconstruction in a dedicated process so that they overlap the next chunk's generation. The audio stream is reconstructed so that each emitted window matches a full offline decode, which removes the discontinuities that a naive sliding decode produces. Together these make the published byte stream identical to an offline render of the same latents while sustaining the real-time generation and sub-second interaction reported in~\cref{tab:latency_master}.

\section{Experiments} \label{sec:experiments} 

In this section, we empirically study MaineCoon.
We evaluate MaineCoon from two complementary perspectives. First, we compare its audio-visual generation quality with representative bidirectional and streaming baselines on SocialVideo-Bench. Second, we measure generation throughput under a unified single-GPU (H100) setting to evaluate whether the model can support real-time streaming deployment. 

\subsection{Experimental Setup} \paragraph{Benchmark} We conduct the main evaluation on SocialVideo-Bench, which contains 700 prompts evenly distributed across seven representative social-video domains. Each sample contains two consecutive 10-second segments, with the conditioning prompt updated at the segment boundary. This protocol evaluates not only visual and audio quality, but also temporal consistency, audio-visual coherence, and the ability to follow a changing condition over a 20-second generation. 

\paragraph{Baselines} We compare MaineCoon with representative audio-visual generation systems from three categories. Bidirectional text-to-audio-video baselines include JavisDiT++~\citep{liu2026javisdit++}, Ovi~\citep{low2025ovi}, JoyAI-Echo~\citep{li2026joyai}, MoVA~\citep{team2026mova}, and LTX-2.3 base/distilled versiosn~\citep{hacohen2026ltx}. Streaming text-and-audio-to-video baselines include LiveAvatar~\citep{huang2025live} and SoulX-FlashTalk~\citep{yusoulx}. For the efficiency comparison, we additionally include streaming text-to-video models, including Causal Forcing~\citep{zhu2026causal}, Helios-Distilled~\citep{yuan2026helios}, and Krea~\citep{krea_realtime_14b}. All available models are evaluated at their officially supported inference configuration on a single GPU as MaineCoon. 

\paragraph{Evaluation metrics} We report nine complementary metrics covering visual quality (\textbf{Vis}), motion quality (\textbf{Mot}), audio quality (\textbf{Aud}), text-video alignment (\textbf{IB-TV}), text-audio alignment (\textbf{IB-TA}), audio-video semantic consistency (\textbf{IB-AV}), audio-visual temporal alignment (\textbf{AV-Al}), Audio-Visual Harmony (\textbf{AVH}), and the Joint Audio-Visual Integrated Score (\textbf{JAVIS}). AVH and JAVIS provide the most comprehensive assessment of joint audio-visual generation quality. We additionally report an aggregate score obtained by averaging the max-normalized values of the nine metrics. 

\paragraph{Efficiency protocol} We measure sampling throughput as the number of newly generated video frames per second. All throughput results are measured for 480P, 20-second generation on a single NVIDIA H100 GPU. The reported FPS result measures the end-to-end sampling process after the input condition has been prepared, which is fair in streaming generation. 

\subsection{Main Results on SocialVideo-Bench} 

\begin{table}[t]
      \centering
      \caption{Main quantitative results of MaineCoon and audio-visual generation models on SocialVideo-Bench. MaineCoon achieves significantly better average scores than popular baseline models. Two most comprehensive metrics, Audio-Visual Harmony (AVH) and Joint Audio-Visual Integrated Score (JAVIS), consistently support the significant advantage of MaineCoon over other streaming and
  bidirectional baseline models.}
      \label{tab:bench600_master}
      \resizebox{\textwidth}{!}{%
      \begin{tabular}{ll|*{9}{c}|cc}
        \toprule
        Type & Model & Vis$\uparrow$ & Mot$\uparrow$ & Aud$\uparrow$ & IB-TV$\uparrow$ & IB-TA$\uparrow$ & IB-AV$\uparrow$ & AV-Al$\uparrow$ & AVH$\uparrow$ &
  JAVIS$\uparrow$ & \textbf{Average}$\uparrow$ \\ 
        \midrule
        \multirow{5}{*}{Bidirectional T2AV}
          & JavisDiT++        & 4.39 & \textbf{2.22} & 4.06 & 0.134 & 0.070 & 0.151 & 0.312 & 0.136 & 0.112 & 0.711 \\
          & Ovi               & 4.44 & 1.89 & 3.76 & \underline{0.138} & 0.079 & 0.191 & \textbf{0.412} & 0.188 & 0.162 & 0.779 \\
          & JoyAI-Echo        & 4.61 & 1.17 & 3.47 & \textbf{0.147} & 0.088 & 0.226 & 0.319 & 0.196 & 0.173 & 0.749 \\
          & MoVA              & \underline{4.66} & 1.68 & 3.69 & 0.133 & 0.105 & 0.258 & \underline{0.359} & 0.245 & 0.216 & 0.842 \\
          & LTX-2.3           & 4.10 & 0.99 & 4.06 & 0.132 & 0.111 & 0.311 & 0.334 & 0.287 & \underline{0.247} & 0.848 \\
        \midrule
        \multirow{2}{*}{Streaming TA2V}
          & LiveAvatar        & 4.60 & 1.46 & \underline{4.13} & 0.131 & 0.120 & \underline{0.316} & 0.326 & \underline{0.291} & 0.246 & 0.892 \\
          & SoulX-FlashTalk   & 4.65 & \underline{1.99} & 4.07 & 0.128 & \underline{0.120} & 0.307 & 0.279 & 0.283 & 0.238 & \underline{0.895} \\
        \midrule
        \multirow{1}{*}{Streaming T2AV}
          & MaineCoon (Ours)  & \textbf{4.71} & 1.62 & \textbf{4.35} & 0.127 & \textbf{0.130} & \textbf{0.318} & 0.334 & \textbf{0.308} & \textbf{0.272} &
  \textbf{0.934} \\
        \bottomrule
      \end{tabular}%
      }
    \end{table}

      \begin{table}[t]
      \centering
      \caption{Latency and model size comparison. MaineCoon with the largest model size can be 7 times faster than other streaming audio-visual generation models and is even faster than a 1.3B streaming video generation model. Sampling throughputs (FPS) are measured for 480P-20s video generation on a single H100 GPU. Sampling throughputs
  can truly reflect new-frame generation speed for streaming generation.
  }
      \label{tab:latency_master}
      \begin{tabular}{ll|*{2}{c}}
        \toprule
        Type & Model & Parameters & FPS  \\
        \midrule
        \multirow{6}{*}{Bidirectional T2AV}
          & JavisDiT++       & 1.8B  & 0.87  \\
          & Ovi              & 11B   & 0.58  \\
          & JoyAI-Echo       & 23B   & 18.0  \\
          & MoVA             & 32B   & 0.26  \\
          & LTX-2.3          & 22B   & 1.40  \\
          & LTX-2.3-Distilled  & 22B   & 20.7  \\
        \midrule
        \multirow{3}{*}{Streaming T2V}
        & Causal-Forcing  & 1.3B   & 19.1   \\
          & Helios-Distilled  & 14B   & 18.2  \\
        & Krea  & 14B   &   6.1   \\
        \midrule
        \multirow{2}{*}{Streaming TA2V}
          & LiveAvatar       & 14B   & 6.7   \\
          & SoulX-FlashTalk  & 14B   & 6.6   \\
        \midrule
        \multirow{1}{*}{Streaming T2AV}
          & MaineCoon (Ours) & 22B   & \textbf{47.5}  \\
        \bottomrule
      \end{tabular}%
    \end{table}

\begin{figure}[t!]
    \centering
    \includegraphics[width=1\linewidth]{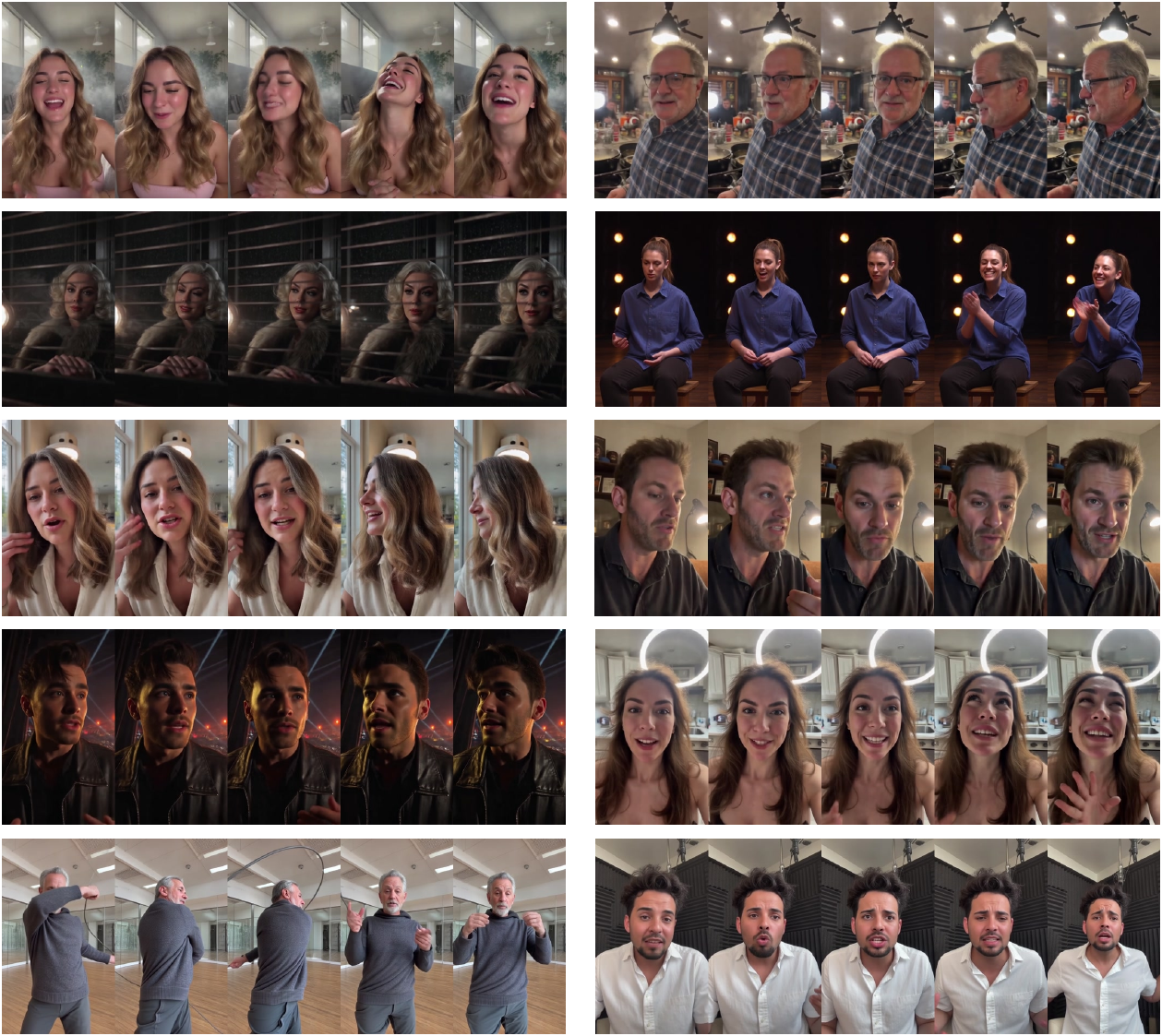}
    \caption{Long audio-isual examples generated by MaineCoon. Our model exhibits strong social-media nativity (e.g., TikTok-style) with highly engaging, authentic look and realistic, finely detailed human portraits. See~\cref{sec:more_cases} for more cases.}
    \label{fig:main_case}
\end{figure}

We present the primary quantitative evaluation on SocialVideo-Bench in~\cref{tab:bench600_master}. MaineCoon achieves the best overall performance among all evaluated systems, attaining an aggregate score of $0.934$, compared with $0.895$ for the strongest baseline. It ranks first on six of the nine individual metrics, including visual quality, audio quality, text-audio alignment, audio-video semantic consistency, Audio-Visual Harmony (AVH), and JAVIS.

The gains are particularly pronounced on the metrics that jointly assess the audio and visual streams. MaineCoon improves AVH from the previous best of $0.291$ to $0.308$, and JAVIS from $0.247$ to $0.272$, corresponding to relative improvements of $5.8\%$ and $10.1\%$, respectively. It also achieves the highest visual and audio quality scores, demonstrating that the improved joint audio-visual consistency is obtained without sacrificing the fidelity of either modality.

Notably, MaineCoon achieves these results under a causal streaming formulation, whereas most competing text-to-audio-video models generate the full sequence bidirectionally. Despite this substantially more demanding generation setting, MaineCoon surpasses both bidirectional audio-visual generators and specialized streaming avatar models in aggregate quality. These results demonstrate that MaineCoon provides a strong balance of visual fidelity, audio quality, semantic alignment, and cross-modal harmony, establishing a new state of the art for streaming text-to-audio-video generation on SocialVideo-Bench.

Additionally, we present representative visual examples from MaineCoon in~\cref{fig:main_case} and throughout~\cref{sec:more_cases}, covering all domains of SocialVideo-Bench including dense speech, two-person interaction, music, emotional performance, dance, and social memes. The generated results exhibit strong \textbf{social-media nativity} (e.g., TikTok-style) with a highly engaging, authentic look and realistic human portraits across diverse scenarios.

\subsection{Streaming Inference Efficiency}
In~\cref{tab:latency_master}, we compare model size and generation throughput. With the training chunk size of 2, MaineCoon already achieves a real-time rate of 31 FPS on a single H100 GPU. More importantly, our model exhibits strong generalization to varying chunk sizes at inference time; by simply increasing the inference chunk size to 6, the throughput substantially improves to 47.5 FPS for 480P video—without any observable degradation in visual quality—delivering the highest throughput among all evaluated systems. This makes it 
4
×
4× faster than the evaluated streaming avatar models. This efficiency is particularly notable because MaineCoon contains 22B parameters and jointly generates both audio and video, confirming that the speedup stems from the native causal architecture, few-step sampler, KV-cache reuse, and agentic streaming inference pipeline, rather than from a smaller backbone. At 47.5 FPS, MaineCoon generates new frames faster than standard real-time playback, leaving additional compute headroom for streaming control and interactive serving.

\section{Position and Outlook: Social World Models}    

In this section, we formally discuss the position and future of social world models in our roadmap.

To establish a functional foundation for a \textit{social world model}, we need to formally transition from modeling rigid physical environments to modeling what we term \textit{social physics}, which are the complex, non-deterministic behavioral laws that govern human-centric interaction. While traditional video world models focus on capturing spatial geometry, fluid dynamics, and object permanence, a social world model assumes that the most critical forces in a scene are psychological, communicative, and interpersonal. 

We may define ``social dynamics'' not as an abstract qualitative concept, but as a high-dimensional joint probability distribution of synchronized multi-modal behavioral tokens. In this section, we try to provide the formal mathematical formulation of these dynamics and deconstruct them into their constitutive temporal, behavioral, and platform-level dimensions.

\paragraph{Formulation of Social World Modeling}
Traditional world models, such as those optimized for robotics or video games, typically formalize environment transitions using a Conditional Markov Decision Process, predicting the next world state $X_{t}$ conditioned on past states and an explicit physical action $U_{t}$:
\begin{equation}
    P(X_{t} \mid X_{<t}, U_{<t}).
\end{equation}
In contrast, a social world model operates over a continuous, multi-modal stream where the primary driver of environmental change is human behavior. Let $V_t$ represent a slice of visual tokens (capturing facial expressions, gestures, and background framing) and $A_t$ represent a slice of acoustic tokens (capturing phonemes, tone, and ambient noise) at time step $t$. Furthermore, let $H_{<t}$ represent the history of multi-modal communicative inputs and the human feelings from the interacting human user.

We formalize the social world model as an autoregressive generation task that computes the joint distribution of visual and acoustic states conditioned on both historical outputs and real-time user perturbations:
\begin{equation}
\label{eq:social_sim}
    P(V_t, A_t, \mid V_{<t}, A_{<t} , H_{<t}).
\end{equation}
By framing social dynamics through this joint autoregressive lens, the model is forced to learn the implicit dependencies between video, audio, user interaction/state.

A crucial distinction must be drawn between the proposed paradigm and prior world modeling literature. Recent scale-driven video models demonstrate an extraordinary capacity to model physical substrates, predicting how a ball bounces, how light reflects off water, or how text renders on a sign. However, these models are fundamentally static regarding human intent; they view humans as moving obstacles or passive textures within a scene. Conversely, a social world model treats the human agent as the primary coordinate system of the generation space. The model optimizes its parameter weights not to preserve pixel-perfect background textures over infinite horizons, but to maximize the likelihood of high-fidelity human expressions, rapid semantic transitions, and emotional resonance.

\paragraph{Next Step Towards Social World Models}
While the current iteration of MaineCoon establishes a critical baseline for real-time human-centric generation, realizing a complete social world model demands a systemic evolution from isolated, half-duplex (turn-based) generation toward highly scalable, persistent multi-user ecosystems. We consider Real-Time Dual-System Full-Duplex Interaction as the next key step toward social world models.

Traditional interactive avatars and conversational models operate on a rigid, half-duplex loop, where the user speaks, the system processes, and the system responds while the user listens passively. This turn-based constraint fundamentally breaks the illusion of social reality; humans do not freeze their expressions or pause their cognitive state while waiting for an API response. 

To achieve true Real-Time Full-Duplex Interaction, a social world model must handle continuous, interlaced multi-modal inputs and outputs (Audio, Video, Text, Gesture) streaming chunk-by-chunk in a non-blocking format. This requires a cognitive bifurcation into a dual-system processing loop:

\begin{itemize}
    \item System 1 (The Fast Cerebellum): A specialized, low-overhead reactive model dedicated entirely to maintaining sub-second response loops. It manages instant temporal synchronization, streaming denoising under strict sink constraints, and real-time reflex generation, such as conversational backchanneling (e.g., nodding or uttering affirmative sounds mid-sentence), fast viseme matching, and immediate gestural feedback.
    \item System 2 (The Planning Brain): A larger, strategic deliberative network operating asynchronously alongside the reactive system. The Planning Brain oversees long-horizon trajectory planning, deep semantic comprehension, long-term persona preference cache management, and structural conversation guidance.
\end{itemize}

The architecture ensures the model can actively observe context and proactively steer conversations without sacrificing the sub-second immediacy vital for human conversational physics. By further including Real-Time Dual-System Full-Duplex Interaction, a complete social world model ensures that every interactive agent within the ecosystem feels alive, immediate, and emotionally resonant through full-duplex tracking.

\paragraph{Outlook and the Road Ahead}
In summary, the transition from simulating rigid physical substrates to modeling the intricate nuances of social dynamics marks a definitive frontier for \textit{artificial social intelligence}. MaineCoon successfully delivers the foundational generative core, proving that streaming, real-time audio-visual autoregression can be executed natively on commodity hardware. It still represents only the initial framework of a broader revolution of social world models. The full realization of social world models ultimately rests on successfully executing this bifurcated, full-duplex cognitive architecture. By permanently retiring the artificial constraints of half-duplex, turn-based loops, and instead processing continuous human feedback through synchronized reactive and deliberative layers, artificial agents will finally mirror the fluid temporal dynamics of organic human communication. As these systems scale from standalone sandbox environments into persistent, high-concurrency multi-user ecosystems, they will fundamentally redefine the digital medium. Grounded in a low-latency, multi-modal interface, future social world models will transition generative AI from a passive content utility into an active, deeply resonant participant in our human social fabric. We believe this direction will ultimately serves as the architectural foundation for next-generation, AI-native social platforms.

\section{Conclusion}
\label{sec:conclusion}

In this work, we have positioned \textit{social world models}, highlighting a critical yet largely overlooked gap in generative AI where human-centric interactive audio-visual simulation have been neglected. A social world model needs to actively observe users, internally simulate social dynamics, and react to human in real time. While traditional world models focus on predicting rigid physical environments or gaming mechanics, they remain decoupled from the fluid nuances of human interaction, such as conversational flow, micro-expressions, and high-engagement pacing. 

The first step towards the paradigm of social world models requires a strong real-time audio-visual generation model. In this work, we presented MaineCoon, a pioneering 22B parameter real-time audio-visual autoregressive generation model. To the best of our knowledge, MaineCoon is not only designed and optimized for social-interactive applications, but also the first audio-visual generation model capable of real-time generation and interaction on a single GPU. 

By introducing a novel multi-stage forcing-free streaming training paradigm, which integrates self-resampling, cross-modal representation alignment, data-domain-aware preference optimization, and ROPD, MaineCoon successfully breaks the computational and latency barriers inherent to real-time multi-modal generation. This architecture delivers sub-second interaction and a record-breaking streaming frame rate of up to 47.5 FPS on a single GPU, dropping generation costs below \$0.001 per second. We also designed the first agentic streaming inference framework that significantly enhance thousand-second-scale long-horizon video generation with agentic cache management and prompt planing. Furthermore, to address the lack of established benchmarks in this domain, we introduced SocialVideo Bench, a comprehensive evaluation framework featuring 9 representative social-interactive metrics. Extensive empirical evaluations demonstrate that MaineCoon significantly outperforms 7 representative open audio-visual generation models, setting a new state-of-the-art (SOTA) standard for both generation quality and inference speed.

Ultimately, the paradigm of social world model marks a fundamental paradigm shift, transitioning generative AI from a passive content-generation tool into an active, responsive participant in the human social dynamics. For future work, we intend to build the active observation module and the internal social simulator module, and joint optimize the three core modules, including the reactive module MaineCoon. By proving that high-quality, ultra-low-cost, real-time social video generation is achievable on a single GPU, this work lays the technical foundation required to realize next-generation, AI-native social platforms.

\section*{Contributions}

\textbf{Core Contributors:} Lichen Bai$^\star$, Tianhao Zhang$^\star$, and Zeke Xie$^\dagger$.\footnote{$^\star$: Equal Contributions; $^\dagger$: Correspondance \& Project Lead.}

\textbf{Contributors:} Shitong Shao, Dingwei Tan, Qiyu Zhong, Zhengpeng Xie, Haopeng Li, Qinghao Huang, Dandan Shen, Tengjiao Ji, Wei Wang, Peicheng Wu, Yuxuan Zhao, Xiangyu Zhu, Welly Luo, and Shurui Yang.


\bibliographystyle{plainnat}
\bibliography{main}


\appendix

\section{More Qualtiative Cases}
\label{sec:more_cases}
In this section, we provide additional qualitative examples generated by MaineCoon across the seven domains of SocialVideo-Bench: dense speech, two-person interaction, music and vocal, emotional performance, dance, and social memes. Each example is generated under the same streaming inference setup described in \cref{sec:inference}. For each domain, we show several consecutive frames to illustrate temporal coherence and audio-visual synchronization. More comprehensive video results can be found in the project page.

\begin{figure}[h!]
    \centering
    \includegraphics[width=0.85\linewidth]{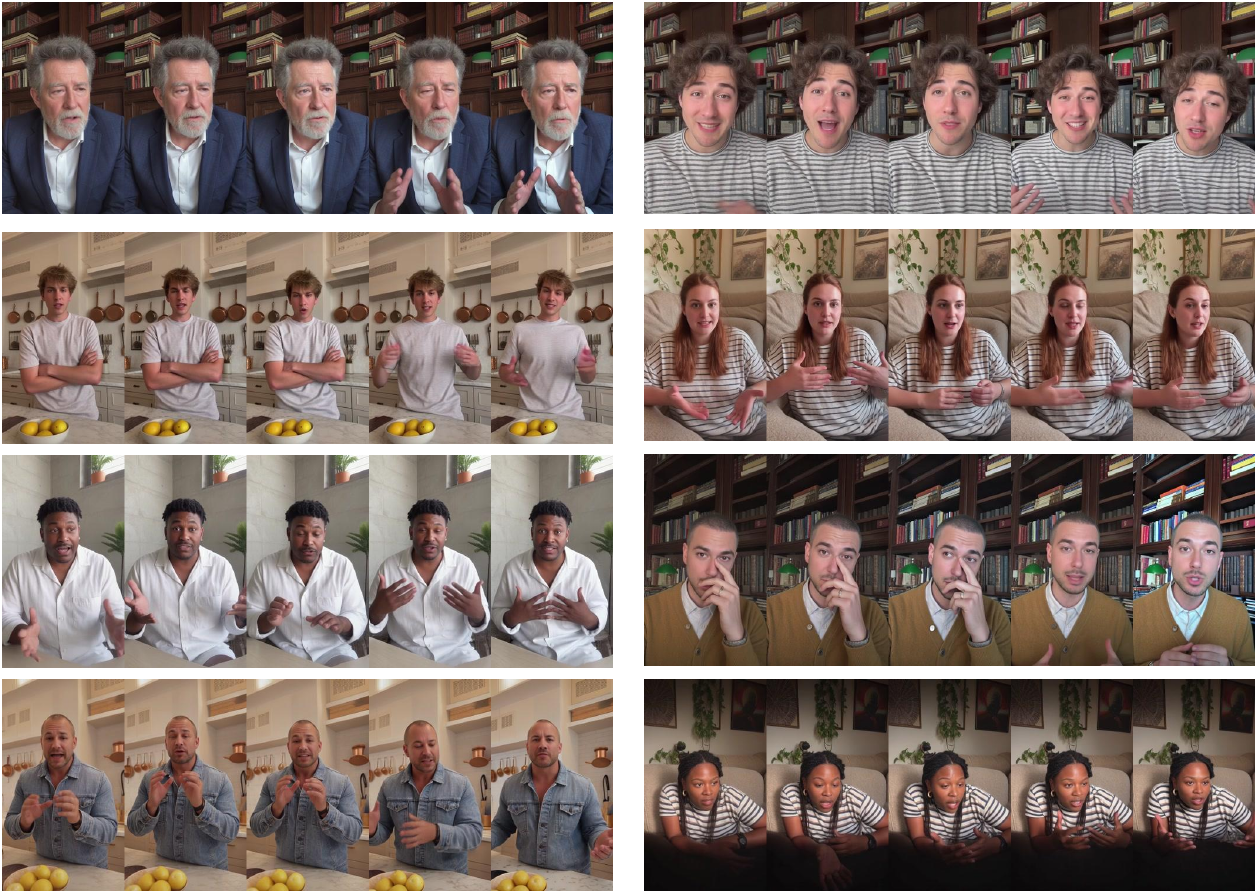}
    \caption{\textbf{Dense Speech}: natural lip synchronization, facial expression, and speech clarity are well preserved over the streaming rollout.}
    \label{fig:dense_speech}
\end{figure}
\vspace{-0.5cm}
\begin{figure}[h!]
    \centering
    \includegraphics[width=0.85\linewidth]{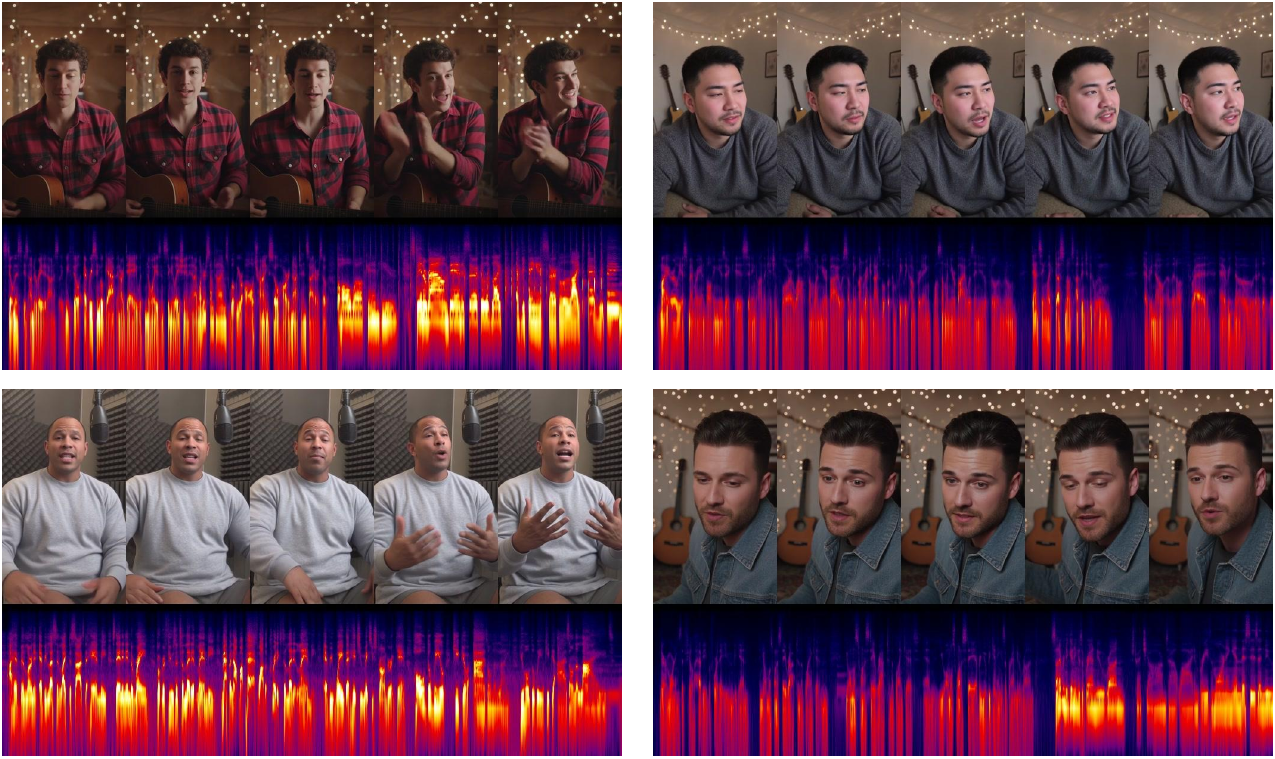}
    \caption{\textbf{Music and Vocal}: vocal timbre and rhythmic motion are consistently maintained across the streaming generation.}
    \label{fig:music}
\end{figure}

\begin{figure}[h!]
    \centering
    \includegraphics[width=0.8\linewidth]{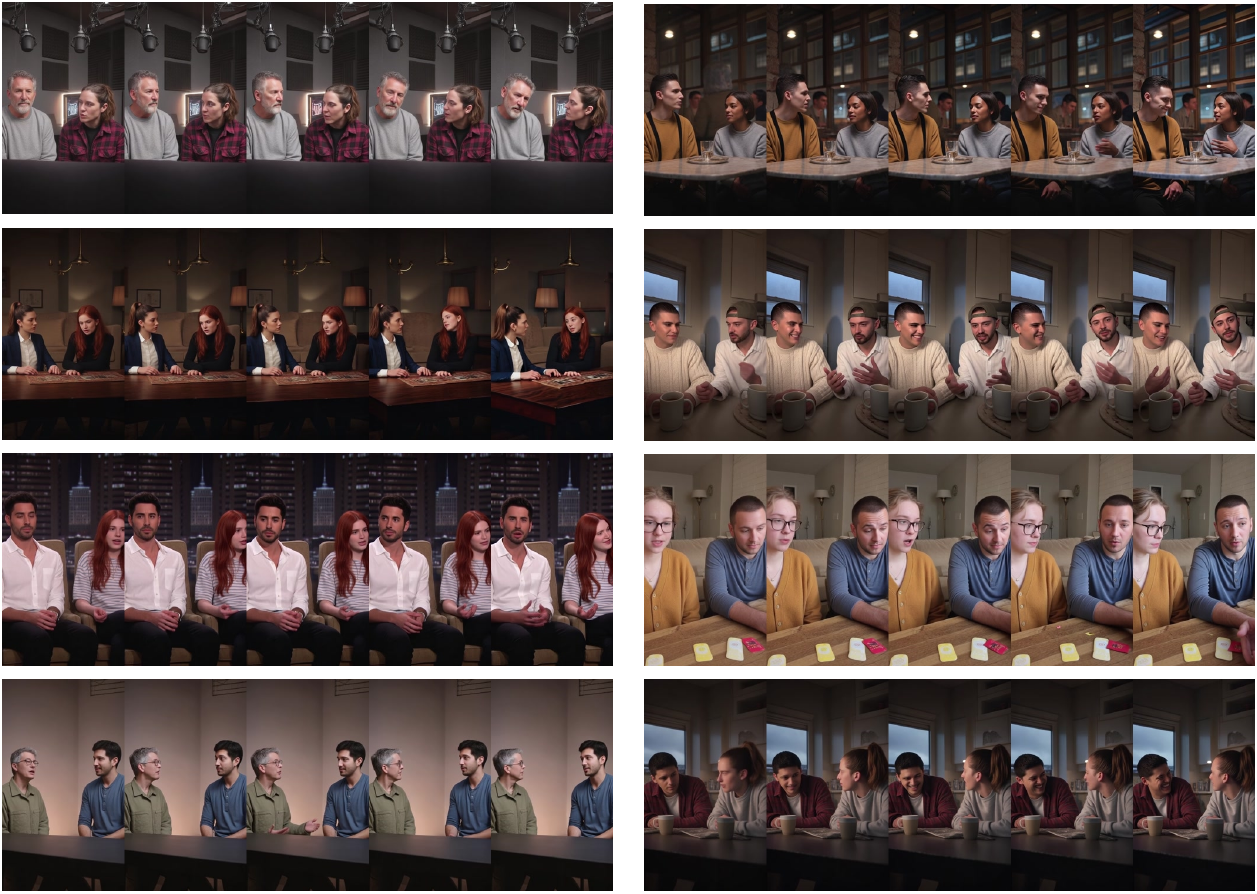}
    \caption{\textbf{Two-Person Interaction}: consistent speaker identity and natural turn-taking are maintained across the streaming rollout, with synchronized lip motion and facial expressions.}
    \label{fig:2_person}
\end{figure}

\begin{figure}[h!]
    \centering
    \includegraphics[width=0.8\linewidth]{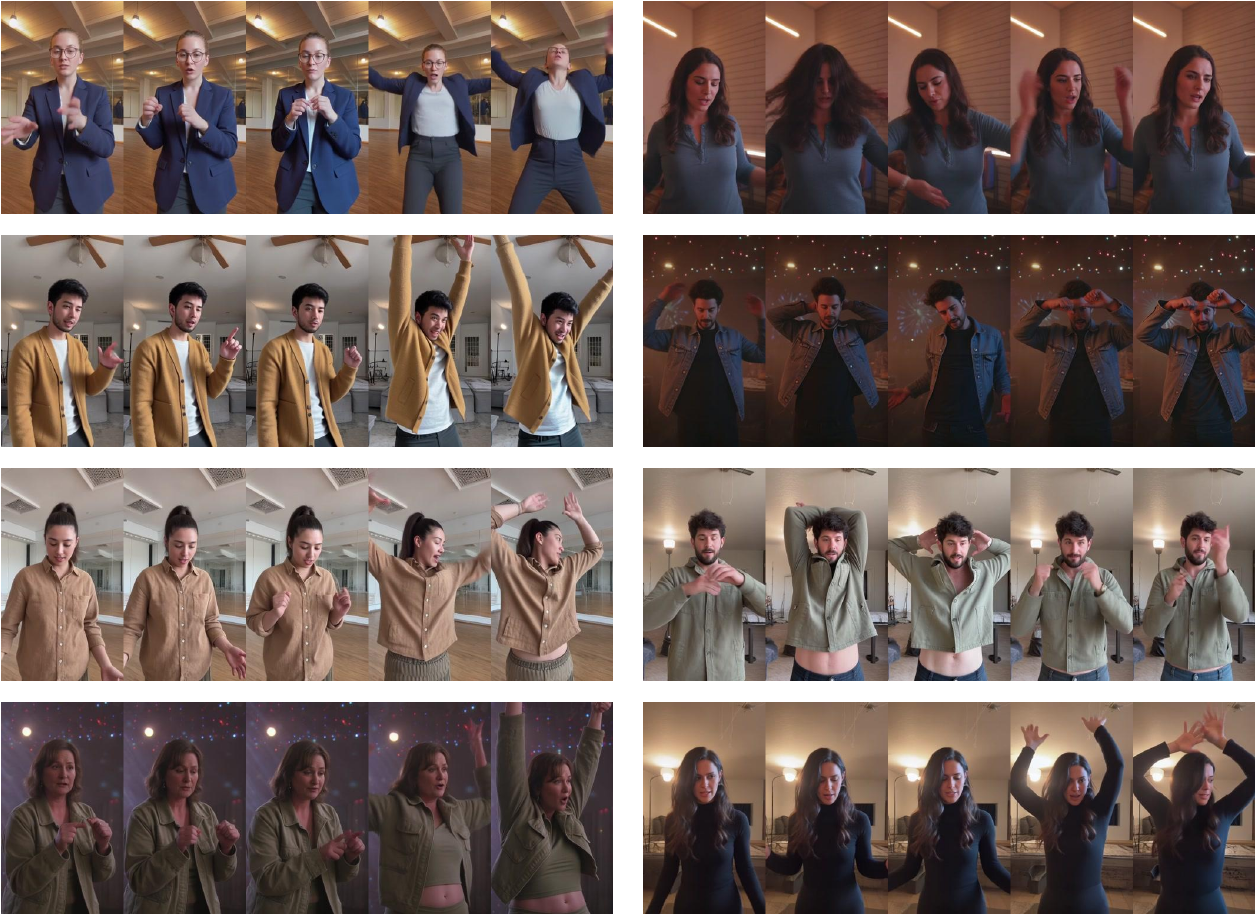}
    \caption{\textbf{Dance and High-Motion}: large body movements remain temporally coherent without tearing or ghosting artifacts.}
    \label{fig:dance}
\end{figure}

\begin{figure}[h!]
    \centering
    \includegraphics[width=0.9\linewidth]{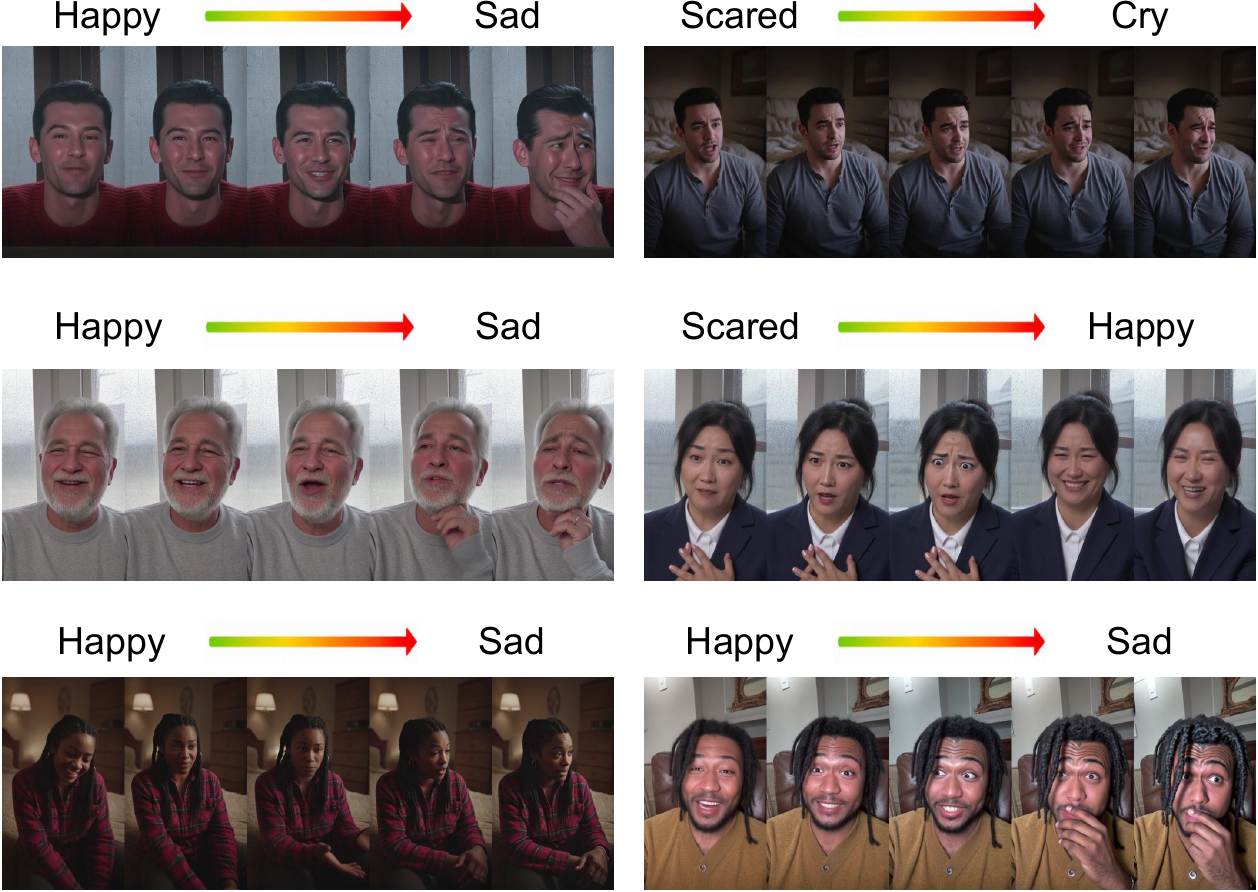}
    \caption{\textbf{Emotional Performance}: expressive facial emotions and prosodic changes are smoothly transitioned across the streaming generation.}
    \label{fig:emotion}
\end{figure}

\begin{figure}[h!]
    \centering
    \includegraphics[width=0.9\linewidth]{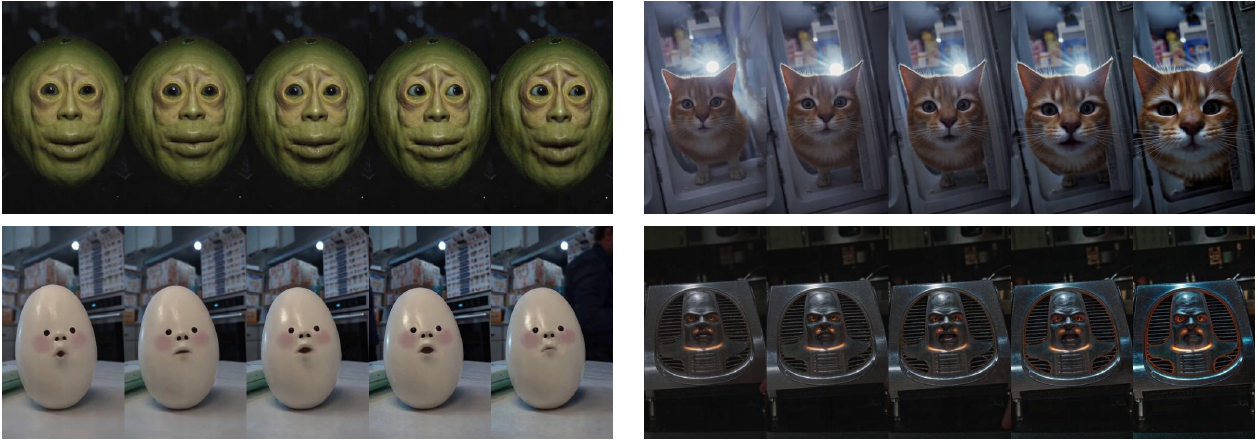}
    \caption{\textbf{Social Memes}: the model captures humorous timing, narrative reversals, and expressive reactions, maintaining engaging social-media style dynamics over the streaming rollout.}
    \label{fig:meme}
\end{figure}

\end{document}